\documentclass{article} 
\usepackage{iclr2026/iclr2026_conference}
\usepackage{times}

\usepackage{amsmath,amsfonts,bm}

\def\eqref#1{equation~\ref{#1}}

\def\1{\bm{1}}

\DeclareMathAlphabet{\mathsfit}{\encodingdefault}{\sfdefault}{m}{sl}
\SetMathAlphabet{\mathsfit}{bold}{\encodingdefault}{\sfdefault}{bx}{n}

\usepackage[utf8]{inputenc} 
\usepackage[T1]{fontenc}    
\usepackage{hyperref}
\usepackage{url}            
\usepackage{booktabs}       
\usepackage{amsfonts}       
\usepackage{nicefrac}       
\usepackage{microtype}      
\usepackage{xcolor}         

\usepackage{graphicx}
\usepackage{multirow}
\usepackage{amsmath}
\usepackage{amssymb}
\usepackage{wrapfig}
\usepackage{sidecap}
\usepackage{float}

\usepackage{pifont}

\DeclareUnicodeCharacter{2212}{-}

\title{To View Transform or Not to View Transform: \\ NeRF-based Pre-training Perspective}

\author{
    Hyeonjun Jeong, Juyeb Shin, Dongsuk Kum \\
    KAIST, Daejeon, Korea\\
    \texttt{\{hyeonjun.jeong, juyebshin, dskum\}@kaist.ac.kr} \\
}

\iclrfinalcopy 
\begin{document}
\lhead{Published as a conference paper at ICLR 2026}

\maketitle

\begin{abstract}
  Neural radiance fields (NeRFs) have emerged as a prominent pre-training paradigm for vision-centric autonomous driving, which enhances 3D geometry and appearance understanding in a fully self-supervised manner.
  To apply NeRF-based pre-training to 3D perception models, recent approaches have simply applied NeRFs to volumetric features obtained from view transformation.
  However, coupling NeRFs with view transformation inherits conflicting priors; view transformation imposes discrete and rigid representations, whereas radiance fields assume continuous and adaptive functions.
  When these opposing assumptions are forced into a single pipeline, the misalignment surfaces as blurry and ambiguous 3D representations that ultimately limit 3D scene understanding.
  Moreover, the NeRF network for pre-training is discarded during downstream tasks, resulting in inefficient utilization of enhanced 3D representations through NeRF.
  In this paper, we propose a novel NeRF-Resembled Point-based 3D detector that can learn continuous 3D representation and thus avoid the misaligned priors from view transformation.
  NeRP3D preserves the pre-trained NeRF network regardless of the tasks, inheriting the principle of continuous 3D representation learning and leading to greater potentials for both scene reconstruction and detection tasks.
  Experiments on nuScenes dataset demonstrate that our proposed approach significantly improves previous state-of-the-art methods, outperforming not only pretext scene reconstruction tasks but also downstream detection tasks.
  \vspace{-10pt}
\end{abstract}

\section{Introduction}

Accurate and fine-grained 3D scene understanding is essential for autonomous driving, supporting critical tasks such as 3D object detection~\citep{reading2021categorical, li2023bevdepth, li2024bevformer}, high-definition (HD) map construction~\citep{liao2023maptr, shin2025instagram}, and occupancy prediction~\citep{tong2023scene, tian2023occ3d}.
To facilitate these open-world perceptions, view transformation backbones~\citep{li2023bevdepth, li2024bevformer, li2022unifying} have drawn great attention, which project multi-view 2D image features into a unified 3D representation on bird’s-eye-view (BEV) or voxel space.
A unified 3D representation, aligning various modalities~\citep{liu2023bevfusion, li2022unifying, yan2023cross, kim2023crn} in a common metric frame, provides a single 3D canvas that can be leveraged across diverse downstream tasks~\citep{hu2023planning, jiang2023vad, weng2024drive}.

In parallel, neural fields, such as NeRFs~\citep{mildenhall2021nerf} and 3DGS~\citep{kerbl20233d}, have emerged as a dominant paradigm for reconstructing 3D representation and synthesizing novel views by learning a continuous field of color and volume density in a self-supervised manner.
Sharing the goal of understanding the 3D environment, recent studies~\citep{yang2024unipad, huang2024selfocc, xu2024gaussianpretrain} proposed combining NeRFs or 3DGS with view transformation, enabling self-supervised pre-training through photometric and depth reconstruction without the need for expensive manual annotations.

\begin{figure*}[t]
  \centering
  \includegraphics[width=\linewidth]{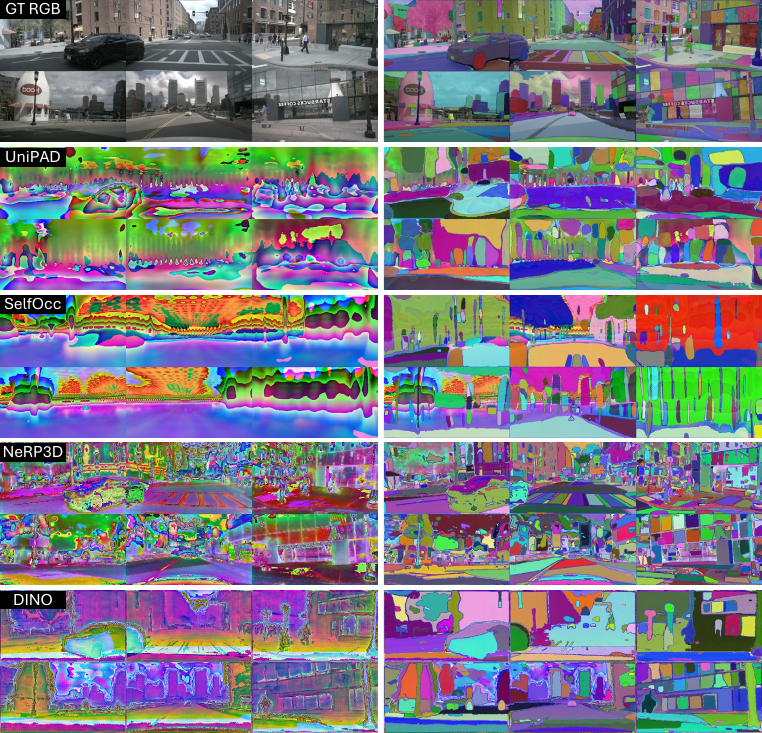}
  \vspace{-10pt}
  \caption{
      Comparison of 2D feature maps (left) and their instance segmentation (right) results using SAM~\citep{kirillov2023sam, ren2024grounded, ravi2024sam2} across different methods.
      All 2D feature maps, except for ground truth RGB (row 1) and DINO~\citep{caron2021emerging, oquab2023dinov2} feature (row 5), are obtained by accumulating 3D point-wise representations along each ray onto the image plane with predicted density. They are extracted directly after radiance field pre-training without any task-specific fine-tuning.
      UniPAD~\citep{yang2024unipad} (row 2) and SelfOcc~\citep{huang2024selfocc} (row 3) produce blurry and inaccurate features that fail to separate nearby or crowded objects, resulting in under-segmented instances. In contrast, NeRP3D (row 4) produces precise and well-localized features with distinct object boundaries without any distillation or fine-tuning from 2D foundation models, comparable to those from DINO features.
      Consequently, we observe the potential for the enhancement of 3D representation to be reflected in the improved segmentation quality.
  }
  \label{fig1}
  \vspace{-10pt}
\end{figure*}

Although both view transformation and NeRFs ultimately aim to reconstruct a 3D representation of the world from 2D signals, they embody conflicting priors. Existing approaches~\citep{yang2024unipad, huang2024selfocc} extract point features for radiance fields by interpolating discretized and fixed voxel features from a view transformation backbone, and then pre-train the backbone through photometric and depth errors rendered from those point features. 
However, this pipeline inevitably leads to NeRF inheriting the discrete and rigid priors of the view transformation, which conflicts with the continuous radiance fields and restricts the fidelity of the reconstructed 3D representation.
Moreover, the pre-trained NeRF is discarded during downstream tasks, preventing effective transfer of NeRF knowledge and limiting the exploitation of enhanced 3D representations from pre-training.
As a result, distinct objects can be collapsed into a single blurry blob, as shown in Fig.~\ref{fig1}.

\begin{figure}[t]
    \centering
    \includegraphics[width=\linewidth]{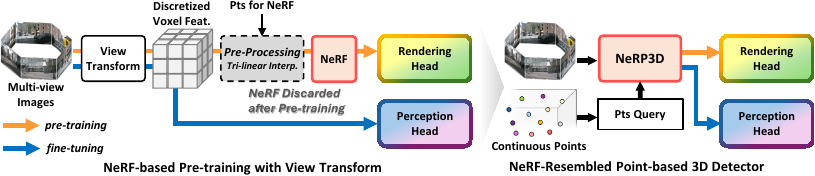}
    \vspace{-15pt}
    \caption{
    Comparison of the previous NeRF-based pre-training methods and our NeRP3D pipeline.
    }
    \label{fig2}
    \vspace{-15pt}
\end{figure}

In this paper, we introduce NeRP3D, a novel NeRF-Resembled Point-based 3D detector that fully inherits the continuous function of neural radiance fields~\citep{mildenhall2021nerf, wang2021neus}, effectively overcoming the inherent discrepancy with view transformation.
Unlike methods relying on rigidly discretized voxel-based representations, NeRP3D directly models 3D scenes as continuous 3D features, geometry, and appearance from any continuous 3D location in a feedforward manner, as illustrated in Fig.~\ref{fig2}.
Experiments on the nuScenes~\citep{caesar2020nuscenes} benchmark demonstrate that our approach significantly improves not only the rendering quality but also the downstream perception tasks for autonomous driving compared to previous approaches that simply incorporate  NeRF-based pre-training into view transformation frameworks.
These findings highlight the importance of aligning the 3D backbone with the pre-training model as well as continuous 3D representation learning in advancing NeRF-based pre-training for enhanced 3D scene understanding.

In summary, our contributions are:
\begin{itemize}
    \item NeRP3D preserves the full knowledge from pre-training, since the NeRF-resembled design makes it effectively inherit and utilize continuous and fine-grained representations for both pretext and downstream tasks.
    \item Regardless of tasks, NeRP3D provides a unified framework allowing for consistent feature extraction with adaptive sampling, ray-wise and uniform spatial sampling, available through our proposed continuous function.
\end{itemize}

\section{Related Work}

\paragraph{Neural Radiance Fields}
Neural radiance fields (NeRFs)~\citep{mildenhall2021nerf} and their variants~\citep{wang2021neus, fridovich2022plenoxels, muller2022instant, barron2022mip, barron2023zip} have established a powerful paradigm for 3D scene reconstruction by learning continuous volumetric functions from posed multi-view images. NeRFs are typically trained in a self-supervised manner, minimizing photometric reconstruction loss across multiple views. 
These prior works have demonstrated their ability to understand and enhance fine 3D geometry and appearance through high-fidelity novel view synthesis and 3D reconstruction.
To move from dense toward sparse image sets, conditioning the radiance fields with image features~\citep{yu2021pixelnerf, chen2021mvsnerf, liu2022neural} shows reliable novel view synthesis results, demonstrating that generic 2D representations can guide NeRF training. Moreover, depth supervision \citep{roessle2022dense, wei2023depth, deng2022depth, wei2021nerfingmvs} is incorporated to understand more accurate geometry.
NeRF's enhanced 3D understanding is increasingly being extended to autonomous driving applications, and NeRP3D aims to fully leverage these capabilities.

\paragraph{Neural Radiance Fields with Autonomous Driving}
The inherent ability of NeRFs~\citep{mildenhall2021nerf, wang2021neus, barron2022mip, barron2023zip} to capture 3D scene structure from multi-view 2D observations in a self-supervised manner has positioned them as a promising foundation for various autonomous driving applications.
For sensor simulation in driving environments, offline scene reconstruction methods~\citep{yang2023unisim, tonderski2024neurad, yang2023emernerf} have demonstrated NeRFs' capability to synthesize realistic camera images, generate scenarios through object manipulation, and decompose static-dynamic scenes. Moreover, DistillNeRF~\citep{wang2024distillnerf} builds upon EmerNeRF~\citep{yang2023emernerf} by extending it into a feed-forward model, while feature distillation from 2D foundation models~\citep{radford2021learning, oquab2023dinov2} further enhances 3D scene understanding.

The most relevant branch of this paper is the integration of NeRFs in pre-training to improve downstream perception tasks. UniPAD~\citep{yang2024unipad} introduces a universal NeRF-based pre-training framework to enhance the 3D object detection downstream task.
Occupancy predictions~\citep{huang2024selfocc, zhang2023occnerf} are also integrated with NeRF, which is optimized through multi-view consistency~\citep{godard2017unsupervised, godard2019digging, zhou2017unsupervised}. 
GaussianPretrain~\citep{xu2024gaussianpretrain} has demonstrated the feasibility of 3D Gaussian Splatting~\citep{kerbl20233d} for pre-training 3D scene representations in driving environments.
However, existing methods~\citep{yang2024unipad, huang2024selfocc, tian2023occ3d, xu2024gaussianpretrain}, which rely on view transformation, have inherent constraints that diminish NeRF's capacity for continuous and fine-grained 3D representation. Moreover, pre-trained NeRF is discarded during downstream tasks, resulting in suboptimal 3D representations enhancement from pre-training.
In contrast, NeRP3D fully inherits pre-trained NeRF knowledge and utilizes continuous and fine-grained representations through its NeRF-resembled design.
\section{Method}

\begin{figure*}[t]
    \centering
    \vspace{-10pt}
    \includegraphics[width=0.8\linewidth]{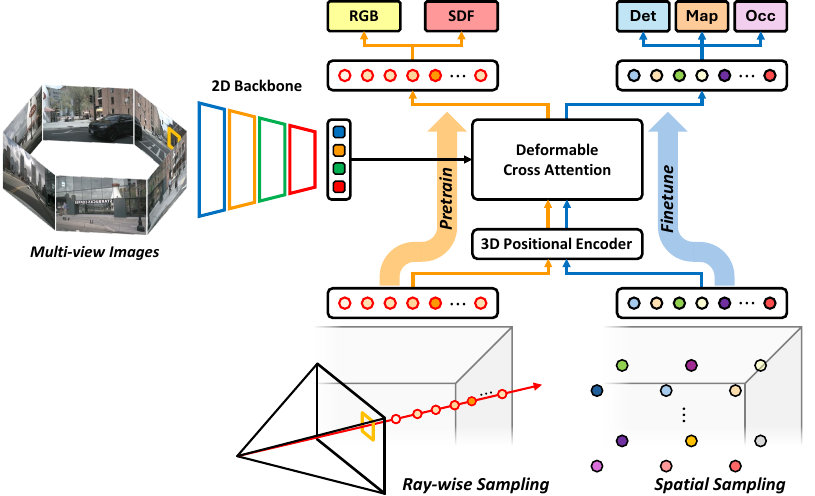}
    \caption{
        Overview of NeRP3D, illustrating both pre-training for rendering (orange) and fine-tuning for downstream (blue) pipelines.
        Through NeRF-resembled design, our method maintains a coherent 3D understanding from scattered points across diverse tasks while accommodating task-specific point sampling strategies, enabling the model to effectively leverage underlying geometric and appearance information while allowing for task-dependent feature specialization.
    }
    \label{fig3}
    \vspace{-10pt}
\end{figure*}

NeRP3D is a simple and effective NeRF-resembled architecture that unifies scene reconstruction and perception tasks from single-timestep multi-view images. As illustrated in Fig.~\ref{fig3}, our framework operates in two distinct stages within a unified architecture, without discarding or adding modules depending on stage or task requirements. This unified architecture enables adaptive exploration of regions of interest tailored to specific processing efficiency, while maintaining a coherent 3D understanding across diverse tasks.

\subsection{Adaptive Sampling \& Representation of Point}
To reconstruct accurate 3D representations from sparse and dynamic multi-view inputs,
NeRP3D directly samples 3D points of interest at arbitrary spatial locations and predicts the representation of sampled points with 2D image features to cope with dynamic driving scenes, without processing voxelized feature grids or any interpolation from them.

NeRP3D first samples 3D points $\mathbf{x}\in\mathbb{R}^3$ using one of two distinct strategies tailored to different processing phases, view-dependent ray-wise sampling and uniform spatial sampling.
For volumetric rendering, we follow the standard NeRF. Specifically, for each pixel in the multi-view images, we define a camera ray $\mathbf{r}_i$ based on its origin $\mathbf{o}_i$ and direction $\mathbf{d}_i$, which are derived from camera intrinsics and extrinsics. Along each ray, we sample a set of points $\{\mathbf{x}_{ij}=\mathbf{o}_i+t_j\mathbf{d}_i\}$ at regular or stratified distances within a defined range $\{t_j|j=1,...,D, t_j<t_{j+1}\}$. These sampled points are then integrated into rendered color and depth along the ray for differentiable volumetric rendering.
In contrast, for downstream tasks, where the goal is to utilize the learned 3D representation for autonomous driving tasks such as 3D object detection or occupancy prediction, we sample points across the scene volume rather than following camera rays. We sample points $\mathbf{x}_{xyz}$ uniformly in 3D space around the vehicle, covering regions relevant to perception tasks.

Despite the difference in sampling methods, all 3D points, whether sampled along camera rays or spatially, are represented in the identical system, ensuring consistency across tasks and sharing a unified spatial understanding. In addition, we parameterize 3D coordinates to account for unbounded environments, inspired by \cite{barron2022mip}:
\begin{equation}
    p(\mathbf{x}')=\left\{
    \begin{array}{ll}
         \alpha\mathbf{x}' &  |\mathbf{x}'|\leq1\\
         \left(1-\frac{(1-\alpha)}{|\mathbf{x}'|}\right)\frac{\mathbf{x}'}{|\mathbf{x}'|} &  |\mathbf{x}'|>1
    \end{array}
    \right.\quad,
\end{equation}
where $p(\cdot)$ denotes a parameterized function that preserves real-scale coordinates for points within the inner range, while distributing distant points proportionally to disparity, including those at infinite distance. $\mathbf{x}'$ denotes normalized $\mathbf{x}$ to the range $[0, 1]$ and $\alpha\in[0, 1]$ denotes the contraction ratio.

After sampling 3D points, a set of 3D points $\{\mathbf{x}\}$ is conditioned with sparse 2D observations to represent 3D dynamic environments in a feed-forward manner. Given \textit{N} multi-view images $\{I_i\}_{i=1}^N$, we feed each image to the image backbone to obtain 2D image features $\mathbf{F}\in\mathbb{R}^{N\times H\times W\times C}$.
Then, to enhance 3D point representations with image-aligned context, we adopt a deformable cross-attention~\citep{zhu2020deformable} with 2D image features $\mathbf{F}$. We first encode each 3D query point $\mathbf{x}$ by $\gamma(\cdot)$ and learn a set of $N_s$ sampling offsets $\{\Delta\pi_s\ |\ s=1,...,N_s\}$ relative to its projected 2D location $\pi(\mathbf{x})$, focusing interaction with relevant image regions. The final representation $\mathbf{z}$ of 3D point $\mathbf{x}$ is defined as:
\begin{equation}
\label{eq2}
    \mathbf{z}=\sum_{h=1}^{N_h}\mathbf{W}_h\sum_{s=1}^{N_s}{\mathbf{A}_{h,s}\mathbf{W}'_s\mathbf{F}(\pi(\mathbf{x})+\Delta\pi_{h,s}(\gamma(p(\mathbf{x}'))))},
\end{equation}
where $\mathbf{N_h}$ denotes the number of heads for multi-head attention. $\mathbf{W_h}\in\mathbb{R}^{C\times (C/N_h)}$ and $\mathbf{W}'_s\in\mathbb{R}^{(C/N_h)\times C)}$ denotes learnable weights and $\mathbf{A}_{hs}$ denotes the attention weights which are normalized as $\sum_s{\mathbf{A}_{h,s}}=1$.
The resulting point embedding $\mathbf{z}$ serves as input to both rendering heads and detection heads described in the following sections.

\subsection{Point-based 3D Scene Reconstruction \& Perception}
\paragraph{Volumetric Rendering}
To support 3D scene understanding for downstream tasks in autonomous driving, we first optimize radiance fields in a self-supervised manner~\citep{yang2024unipad}, using the signed distance function (SDF) and RGB reconstruction to represent 3D geometry and appearance. Given a set of sampled points along each ray and its embedded features $\{\mathbf{z}_{ij}\}$, RGB color values of 3D points $\mathbf{x}_j$ are predicted by $c_j=\phi_{rgb}(\mathbf{z}_j,\mathbf{d}_i)$, and its signed distance $s_j$ extracted by signed distance function $\phi_{sdf}(\mathbf{z}_j)$ is transformed into opacity $\alpha_j$ derived with:
\begin{equation}
    \alpha_j=\max{\left(\frac{\Phi_\omega(\phi_{sdf}(\mathbf{z_j}))-\Phi_\omega(\phi_{sdf}(\mathbf{z}_{j+1}))}{\Phi_\omega(\phi_{sdf}(\mathbf{z}_j))},0\right)},
\end{equation}
where $\Phi_\omega(x)=(1+e^{-\omega x})^{-1}$ is the sigmoid function with a learnable parameter $\omega$. Then, the unbiased and occlusion-aware weights~\citep{wang2021neus} $w_j=T_j\alpha_j$ is computed from $\alpha_j$, where $T_j=\prod_{k=1}^{j-1}(1-\alpha_k)$ is the accumulated transmittance. The final color and depth values are computed by accumulating the contributions of 3D points sampled along ray $\mathbf{r}_i$, weighted by the probability distribution $\{w_j\}$:
\begin{equation}
    \hat{\mathbf{C}}(\mathbf{r}_i)=\sum_{j=1}^D{w_j\mathbf{c}_j},\quad \hat{D}(\mathbf{r}_i)=\sum_{j=1}^D{w_jt_j},
\end{equation}
where $\hat{\mathbf{C}}(\mathbf{r}_i)$ and $\hat{D}(\mathbf{r}_i)$ denote the predicted color and depth corresponding to the ray $\mathbf{r}_i$, respectively.

To optimize the neural radiance field, we employ a combination of RGB reconstruction, depth supervision, and multi-view consistency losses.
We adopt the standard volumetric rendering loss from NeRFs, comparing the rendered color $\hat{\mathbf{C}}(\mathbf{r}_i)$ against the ground truth pixel color $\mathbf{C}(\mathbf{r}_i)$ for sampled rays $\mathcal{R}=\{\mathbf{r}_i\}$.
To further constrain the 3D geometry, we leverage explicit depth supervision~\citep{deng2022depth, yang2024unipad} for $\mathbf{r}_i$ against LiDAR measurements $D_{lidar}(\mathbf{r}_i)$ where available.
Furthermore, while LiDAR provides direct supervision, it suffers from sparse scan patterns and cannot capture regions such as the sky, transparent surfaces (\textit{e.g.}, windows), or distant backgrounds where depth is undefined or unprojectable. To address this without additional annotations~\citep{yang2023emernerf} or distillation from 2D foundation models~\citep{oquab2023dinov2, kirillov2023sam}, we further enforce multi-view consistency~\citep{godard2019digging, cao2023scenerf} by minimizing the discrepancy in predicted depth distributions across different views as:
\begin{equation}
    \mathcal{L}_{reproj}=\frac{1}{|\mathcal{R}|}\sum_{\mathbf{r}_i\in\mathcal{R}}\sum_{\mathbf{x}_j\in\mathbf{r}_i}{w_j|I_t(\mathbf{r}_i) - I_s(\pi_{s}(\mathbf{x}_j))|},
\end{equation}
where $I_t(\mathbf{r}_i)$ denotes the color value of a pixel in a target or current image $I_t$ corresponding to the ray $\mathbf{r}_i$.
$\pi_s(\mathbf{x})$ denotes the projection matrix from 3D points to 2D pixels on a source image $I_s$, such as a previous $I_{t-1}$ or future image $I_{t+1}$.
Consequently, the sampled 3D point $\mathbf{x}_j=\mathbf{o}_i+t_j\mathbf{d}_i$ along the ray $\mathbf{r}_i$ is projected on the source image, and the corresponding pixel color $I_s(\pi_{s}(\mathbf{x}_j))$ is compared with $I_t(\mathbf{r}_i)$ in weighted sum $\{w_j\}$.
The overall loss for pre-training consists of RGB reconstruction loss, depth supervision loss, and reprojection loss:
\begin{equation}    
    \mathcal{L}_{pretrain} =
    \lambda_{rgb}\mathcal{L}_{rgb} + \lambda_{depth}\mathcal{L}_{depth} + \lambda_{reproj}\mathcal{L}_{reproj}
\end{equation}
where $\lambda_{rgb}$, $\lambda_{depth}$, and $\lambda_{reproj}$ are the loss scale factors for each pre-training loss. $\mathcal{L}_{rgb}$ is RGB reconstruction loss and $\mathcal{L}_{depth}$ is depth estimation loss directly supervised by LiDAR measurements.

\paragraph{Open-World Perception}
Unlike view-dependent volumetric rendering, perception tasks require comprehensive spatial coverage of the vehicle's surroundings. All we need to do with NeRP3D is scatter the points $\{\mathbf{x}\}\in\mathbb{R}^{N\times 3}$ throughout the space and reshape the resulting representations $\{\mathbf{z}\}\in\mathbb{R}^{N\times C}$ from Eq.~\ref{eq2} to be compatible with task-specific detection heads, for example, $\{\mathbf{z}\}\in\mathbb{R}^{(X\times Y\times Z)\times C}$ for occupancy prediction.
This straightforward adaptation maintains the enhanced geometric and appearance information learned during pre-training while enabling seamless integration with established perception architectures.

\section{Experiments}

We demonstrate NeRP3D on the nuScenese~\citep{caesar2020nuscenes} dataset against the \textit{state-of-the-art} NeRF-based pre-training approaches as well as comparable methods. Our experiments are designed to assess both pre-trained 3D representations by scene reconstruction and the effectiveness of finetuning for downstream tasks.

\subsection{Dataset}
We conduct experiments using the nuScenes dataset~\citep{caesar2020nuscenes}, which provides 700, 150, and 150 scenes for training, validation, and testing, respectively. We follow this data split for both the pretext and downstream tasks. Each scene provides 6 RGB camera images that cover a full 360° field of view, along with a 32-beam LiDAR point cloud and 3D radar point cloud data. The key samples are annotated at 2 Hz and support multiple tasks for autonomous driving, including 3D object detection, HD map construction, and segmentation.
Recently, the annotations for occupancy prediction have been made available through Occ3D~\citep{tian2023occ3d} and SurroundOcc~\citep{wei2023surroundocc}, providing dense 3D semantic occupancy labels. In our experiments, we adopt the Occ3D benchmark for the occupancy prediction.

Moreover, to evaluate generalization across different data distributions and sensor configurations, we additionally utilize Argoverse 2 (AV2)~\citep{wilson2023argoverse} dataset. AV2 provides 1,000 driving sequences with a distinct sensor suite comprising seven high-resolution ring cameras ($2048\times1550$) covering a $360^{\circ}$ field of view and two 32-beam LiDARs.
This setup introduces significant domain shifts in environmental statistics and sensor layouts compared to nuScenes~\citep{caesar2020nuscenes}. This distinct setup serves to assess the model's robustness to domain changes and its data efficiency under limited supervision. For our experiments, we resized the input images to $800 \times 450$ and utilized only a $1/4$ subset of the training data.

\subsection{Evaluation Metrics}
\label{metric}
We evaluate performance across two pretext scene reconstruction tasks and three downstream 3D perception tasks by following standard evaluation protocols for each task to ensure comparability with existing methods.

\paragraph{Scene Reconstruction Tasks}
We compare scene reconstruction quality by generating rendered RGB and depth images 1:4 the size of the input images.
RGB reconstructed images are evaluated for all rendered pixels by Peak Signal-to-Noise Ratio (PSNR), Structural Similarity Index Measure (SSIM), and Learned Perceptual Image Patch Similarity (LPIPS), following standard NeRF evaluation protocols. For depth estimation, we report relative errors (AbsRel \& SqRel), root mean squared error (RMSE \& RMSE log), and accuracy under threshold $\delta$ metrics up to $80m$, only for pixels where the lidar point cloud with 20 sweeps is projected.

\paragraph{Downstream Tasks}
We evaluate the performance of 3D object detection using the mean Average Precision (mAP) and nuScenes Detection Score (NDS) under the standard nuScenes evaluation protocol. The perception range for object detection is set to $[-51.2m,\ 51.2m]$ along both the X and Y axes.
For vectorized HD map construction, we follow the nuScenes map annotation benchmark and report mAP under \textit{Chamfer} distance thresholds ($\tau\in\{0.5, 1.0, 1.5\}$). The evaluation range is set to $[−15.0m,\ 15.0m]$ for the X axis and $[−30.0m,\ 30.0m]$ for the Y axis.
Occupancy prediction aims to predict the semantic classes of $0.4m \times 0.4m \times 0.4m$ voxels covering $[−40m,\ 40m]$ in both the X and Y axes and $[−1.0m,\ 5.4m]$ along the Z axis. The prediction result is evaluated using mean Intersection over Union (mIoU) across 17 semantic classes.

\begin{table}[tb]
\centering
\vspace{-10pt}
\caption{
    \textbf{Downstream detection performance}
}
    \begin{minipage}{0.45\textwidth}
    \centering
    (a) \textbf{3D object detection} \\
    \resizebox{\linewidth}{!}{%
        \begin{tabular}{l|c|cc}
        \toprule
        Method & Pre-train & NDS$\uparrow$ & mAP$\uparrow$ \\
        \midrule
        UVTR-C      & ImageNet & 44.1 & 37.2 \\ 
        BEVFormerV2 & ImageNet & 46.7 & 39.6 \\ 
        \midrule
        TPVFormer$^\dagger$       & SelfOcc & 33.5 & 31.0 \\ 
        UVTR-C$^\dagger$          & UniPAD  & 37.1 & 33.7 \\ 
        \textbf{NeRP3D}$^\dagger$ & Ours    & \textbf{39.2} & \textbf{35.8} \\ 
        \midrule
        UVTR-C                    & UniPAD  & 45.5 & 41.6 \\ 
        \textbf{NeRP3D}           & Ours    & \textbf{47.3} & \textbf{42.8} \\ 
		\bottomrule
        \end{tabular}%
    }
    \end{minipage}
    \hspace{20pt}
    \begin{minipage}{0.22\textwidth}
    \centering
    (b) \textbf{Occ prediction} \\
    \resizebox{\linewidth}{!}{%
        \begin{tabular}{l|c}
        \toprule
        Method & mIoU \\
        \midrule
        BEVDet          & 19.38 \\ 
        BEVFormer       & 26.88 \\ 
        TPVFormer       & 27.83 \\ 
        CTF-Occ         & 28.53 \\ 
        \midrule
        SelfOcc         & 29.65 \\ 
        UniPAD          & 34.05 \\ 
        \textbf{NeRP3D} & \textbf{35.49} \\ 
        \bottomrule
        \end{tabular}
    }
    \vspace{5pt}
    \end{minipage}

    \vspace{10pt}
    \begin{minipage}{0.45\textwidth}
    \centering
    (c) \textbf{HD map construction} \\
    \resizebox{\linewidth}{!}{%
        \begin{tabular}{l|c|c|c}
        \toprule
        Method  & Pre-train & Epochs & mAP \\ 
        \toprule
        HDMapNet        & ImageNet &  30 & 23.0 \\ 
        VectorMapNet    & ImageNet & 110 & 40.9 \\ 
        MapTR-tiny      & ImageNet &  24 & 49.9 \\ 
        \midrule
        TPVFormer       & SelfOcc  &  24 & 53.9 \\ 
        UVTR-C          & UniPAD   &  24 & 57.8 \\ 
        \textbf{NeRP3D} & Ours     &  24 & \textbf{59.1} \\ 
        \bottomrule
        \end{tabular}
    }
    \end{minipage}
\label{tab:downstream}
\vspace{-10pt}
\end{table}

\subsection{Implementation Details}
To ensure fair comparison with prior works~\citep{yang2024unipad, huang2024selfocc}, we adopt identical pre-training architectures and detection heads.
We leverage NeuS~\citep{wang2021neus} for radiance field pre-training, following previous studies.
Furthermore, we conduct downstream tasks based on UVTR-C~\citep{li2022unifying}, MapTR~\citep{liao2023maptr}, and Occ3D (CTF-Occ)~\citep{tian2023occ3d} for 3D object detection, HD map construction, and occupancy prediction, respectively. Class-balanced sampling (CBGS) or specialized data augmentations are not applied for finetuning, and all downstream tasks are trained and evaluated using single-timestep images only, without temporal information or frame stacking.

Our implementation builds upon the MMDetection3D~\citep{mmdet3d2020} framework, and training is conducted on 4 NVIDIA A6000 GPUs.
The input image resolution varies by tasks, set to $1600\times900$ for object detection and $800\times450$ for rendering, HD map construction, and occupancy prediction.
We both pre-train and fine-tune the model for 24 epochs using the AdamW optimizer, with an initial learning rate of 2e-4 and a weight decay of 0.01. The loss scale factors are set to $\lambda_{rgb}=\lambda_{depth}=\lambda_{reproj}=10$. Unless otherwise specified, we fine-tune the models on a 1/2 subset for 12 epochs with $800\times450$ images in ablation studies.

\subsection{Main Results}
\label{main_results}

\begin{figure*}[!t]
    \centering
    \includegraphics[width=1.0\linewidth]{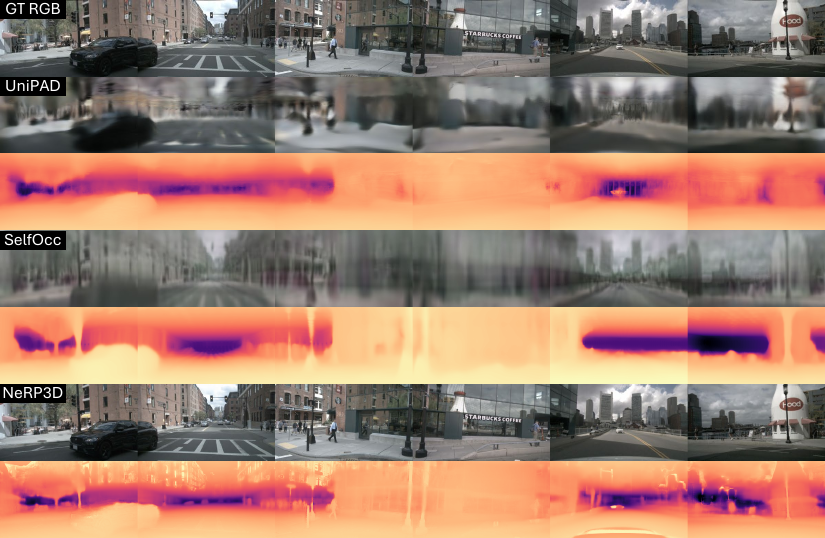}
    \vspace{-20pt}
    \caption{
        Qualitative comparison on rendered RGB \& depth.
        NeRP3D outperforms \textit{state-of-the-art} methods on both RGB and depth reconstruction. Our approach maintains high fidelity in urban scenes without any blur and pattern artifacts. For depth estimation, NeRP3D distinguishes individual people in crowded areas rather than merging them into indistinct blobs, and precisely captures thin structures such as poles that are often missed or reconstructed as thick structures by competing methods.
    }
    \label{fig4}
    \vspace{-10pt}
\end{figure*}

\paragraph{3D Object Detection}
We compare NeRP3D with previous 3D object detection approaches~\citep{li2024bevformer, li2022unifying, liu2022petr, shu20233dppe, yang2023bevformer, yan2023cross} on the nuScenes \textit{val} set.
To compare with previous NeRF-based pre-training methods on detection, we follow the fine-tuning framework of UniPAD~\citep{yang2024unipad} and also reproduce the results of both UVTR-C (UniPAD)~\citep{li2022unifying, yang2024unipad} and TPVFormer (SelfOcc)~\citep{huang2023tri, huang2024selfocc} by replacing the NeRF network for pre-training with UVTR's object detection head.
$\dagger$ in Tab~.\ref{tab:downstream} (a) denotes the result evaluated on input resolutions of $800\times450$.
Compared to the \textit{state-of-the-art} NeRF-based self-supervision methods, our method outperforms 1.8 mAP and 2.1 NDS on $800\times450$ 1.2 mAP and 1.8 NDS on $1600\times900$ over UniPAD, as shown in Tab.~\ref{tab:downstream} (a).
This improvement stems from NeRP3D's ability to learn fine-grained 3D representations, which enables more precise localization of bounding boxes and better separation of nearby objects, as qualitatively suggested by the detailed features in Fig.~\ref{fig1} and sharp reconstructions in Fig.~\ref{fig4}.

\begin{table}[tb]
\centering
\caption{
    \textbf{Pretext scene reconstruction performance}
}
\vspace{10pt}
    \begin{minipage}{0.55\textwidth}
    \centering
    (a) \textbf{Depth estimation} \\
    \resizebox{\linewidth}{!}{%
        \begin{tabular}{l|cccc}
        \toprule
        Method          & Abs Rel$\downarrow$ & Sq Rel$\downarrow$ & RMSE$\downarrow$ & RMSE log$\downarrow$ \\ 
        \midrule
        SelfOcc         & 0.311 & 3.808 & 8.503 & 0.391 \\ 
        SelfOcc$^\ast$  & \underline{0.215} & 2.743 & \textbf{6.706} & \textbf{0.316} \\ 
        UniPAD          & 0.218 & \underline{2.512} & 7.937 & 0.356 \\ 
        \textbf{NeRP3D} & \textbf{0.183} & \textbf{2.274} & \underline{7.884} & \underline{0.353} \\ 
        \bottomrule
        \end{tabular}
    }
    \end{minipage}
    \hspace{3pt}
    \begin{minipage}{0.4\textwidth}
    \centering
    (b) \textbf{RGB reconstruction} \\
    \resizebox{\linewidth}{!}{%
        \begin{tabular}{l|ccc}
        \toprule
        Method          & PSNR$\uparrow$  & SSIM$\uparrow$  & LPIPS$\downarrow$ \\
        \midrule
        SelfOcc         & 18.82           & 0.536           & 0.657             \\
        UniPAD          & 21.14           & 0.549           & 0.634             \\
        \textbf{NeRP3D} & \textbf{33.42}  & \textbf{0.969}  & \textbf{0.070}    \\
        \bottomrule
        \end{tabular}
    }
    \vspace{6pt}
    \end{minipage}
\label{tab:pretext}
\end{table}

\paragraph{Occupancy Prediction}
In Tab.~\ref{tab:downstream} (b), we evaluate the performance of our method on Occ3D-nuScenes for 3D occupancy prediction.
Similar to 3D object detection, we fine-tune the backbones with the same occupancy prediction head~\citep{tian2023occ3d} after pre-training.
Our approach inherits NeRF’s strength in modeling fine-grained representations, leading to improved mIoU and consistent gains over UniPAD and SelfOcc. As a result, our NeRP3D outperforms UniPAD and SelfOcc by 2.8 and 9.2 mIoUs, respectively.
The continuous and high-fidelity representations learned by NeRP3D are particularly beneficial for this dense prediction task, enabling the model to accurately discern object boundaries and capture intricate geometric details often missed by other methods.

\paragraph{HD Map Reconstruction}
We evaluate the accuracy of HD map construction to assess each method’s capability for understanding static driving environments, particularly in detecting road boundaries, dividers, and pedestrian crossings.
To facilitate this task, we commonly utilized the detection head of MapTR~\citep{liao2023maptr} for fair comparison.
As shown in Tab.~\ref{tab:downstream} (c), our method achieves improved mAP compared to both UniPAD and SelfOcc, with gains of 1.3 and 5.2 mAP, respectively.
HD map reconstruction is particularly challenging as it requires a nuanced semantic understanding to differentiate map elements like pedestrian crossings that are geometrically coplanar with the drivable surface. As visually evidenced in Fig.~\ref{fig1}, the feature representations from NeRP3D make these elements distinctly separable, which is critical for precise map construction.

\paragraph{RGB \& Depth Reconstruction}
To validate the effectiveness of the pre-training, the performance of NeRP3D on the pretext tasks is also compared with the previous NeRF-based pre-training methods~\citep{yang2024unipad, huang2024selfocc} on the nuScenes \textit{val} set.
As shown in Tab.~\ref {tab:pretext}, NeRP3D achieves remarkable enhancements in both depth estimation and RGB reconstruction.
More specifically, the qualitative depth maps in Fig.~\ref{fig4} consistently demonstrate that our method yields more accurate and detailed depth estimations, particularly in complex regions, whereas UniPAD and SelfOcc struggle to resolve fine structures and depth discontinuities.
For RGB reconstruction, UniPAD generates blurry and imprecise reconstructions lacking detailed textures, while SelfOcc produces grayish images with unidentified vertical patterns. In contrast, our approach reconstructs sharper images with rich colors, closely matching the ground truth without introducing patterned signals.

\paragraph{Generalization}
To assess the robustness of our method against domain shifts and varying sensor configurations, we conducted cross-dataset transfer experiments using Argoverse 2 (AV2)~\citep{wilson2023argoverse} for pre-training and nuScenes for evaluation. AV2 possesses distinct camera geometries and environmental statistics compared to nuScenes, serving as a rigorous testbed for generalization.

We first evaluated zero-shot scene reconstruction by directly applying the AV2~\citep{wilson2023argoverse} pre-trained weights to nuScenes~\citep{caesar2020nuscenes} without any fine-tuning. As presented in Tab.~\ref{tab:domain}, NeRP3D demonstrates remarkable robustness, achieving an Abs Rel of 0.626 and PSNR of 28.24, significantly outperforming UniPAD (Abs Rel 0.985, PSNR 18.67).
While the view transformation method~\citep{yang2024unipad} suffers from severe degradation due to its dependency on fixed grid priors aligned with specific sensor layouts, NeRP3D’s continuous point-based architecture effectively adapts to new sensor geometries. Qualitative results in Fig.~\ref{fig10} of Appendix further visualize this, showing that NeRP3D preserves structural details while the view transformation method produces blurry artifacts.

Moreover, we evaluated the transferability for 3D object detection. When pre-trained on AV2~\citep{wilson2023argoverse} and fine-tuned on nuScenes~\citep{caesar2020nuscenes}, NeRP3D achieved 27.46 mAP, surpassing UniPAD~\citep{yang2024unipad} (26.29 mAP) by a significant margin. This confirms that NeRP3D learns universal geometric representations that are not overfitted to specific sensor configurations or dataset distributions but are effectively transferable across domains.

Overall, these results demonstrate that our approach effectively leverages the inherent advantages of continuous and fine-grained representations derived from NeRF.
NeRP3D not only significantly benefits pretext scene reconstruction tasks and downstream detection tasks but also ensures robust generalization across different sensor configurations and data distributions for autonomous driving.
More comprehensive comparison and quantitative analysis of the experimental results are provided by Tab.~\ref{tab:det}$-$\ref{tab:rgb} in Appendix~\ref{sup_a} and \ref{sup_b}.

\begin{table}[tb]    
    \vspace{-10pt}
    \centering
    \caption{\textcolor{black}{
        \textbf{Zero-shot scene reconstruction performance (Argoverse 2 $\rightarrow$ nuScenes)}
    }}
    \resizebox{0.85\linewidth}{!}{%
        \begin{tabular}{l|cccc|ccc}
        \toprule
        Method          & Abs Rel$\downarrow$ & Sq Rel$\downarrow$ & RMSE$\downarrow$ & RMSE log$\downarrow$ 
                        & PSNR$\uparrow$      & SSIM$\uparrow$     & LPIPS$\downarrow$ \\
        \midrule
        UniPAD          & 0.985               & 11.767             & 14.963           & 4.390
                        & 18.668              & 0.432              & 0.577 \\
        \textbf{NeRP3D} & \textbf{0.626}      & \textbf{6.251}     & \textbf{10.728}  & \textbf{0.921}
                        & \textbf{28.238}     & \textbf{0.905}     & \textbf{0.111} \\
        \bottomrule
        \end{tabular}
    }
\label{tab:domain}
\vspace{-10pt}
\end{table}

\subsection{Ablation Studies}
We conduct comprehensive ablation studies to analyze different model variants and evaluate their impact.
Ablation results are reported in Appendix~\ref{sup_c} and summarized in the following sections.

\paragraph{Cross-Task Generalization}
We further investigate whether the learned 3D representation remains valid across different task objectives. By performing volumetric rendering using the backbone fine-tuned for occupancy prediction, we observe that NeRP3D successfully retains structural details, whereas view-transformation methods suffer from catastrophic forgetting, collapsing into mean regression. This confirms that NeRP3D learns a task-agnostic continuous representation that preserves geometric fidelity regardless of downstream optimization pressure.

\paragraph{Adaptability}
View transformation is dependent on the range and voxel size, leading to severe performance degradation if the voxel-related parameters are changed against pre-training. In contrast, NeRP3D aims for a continuous representation without voxel-related parameters, and variations only correspond to simple changes in the range of interest.

\paragraph{Effectivness}
We analyze the effectiveness of NeRP3D in reducing the reliance on annotations by comparing previous works, ranging from the full dataset to a 1/8 subset. Consequently, NeRP3D maintains strong detection performance even with significantly reduced supervision, indicating the robustness of its NeRF-based pre-training.

\paragraph{Multi-view Consistency}
LiDAR-based supervision ensures more consistent depth estimation accuracy. However, we found that the sparsity and scan patterns of LiDAR are ultimately insufficient for reconstructing dense 3D geometry. To address LiDAR's sparsity and patterns, we not only rely on LiDAR supervision but also consider multi-view consistency and our sampling strategy tailored to this approach.

\paragraph{Design Validation}
We verify the necessity of our architectural choices through comprehensive comparisons. First, applying NeRF pre-training to existing point-based detectors~\citep{liu2023petrv2} fails to transfer knowledge due to query mismatch, confirming the importance of our unified design. Second, comparisons between SDF (NeuS~\citep{wang2021neus}) and density (standard NeRF~\citep{mildenhall2021nerf}) priors validate that SDF enforces clearer object boundaries beneficial for perception. Finally, we demonstrate that deformable attention outperforms standard attention by providing a necessary locality inductive bias, ensuring that the points remain faithful to their local spatial context.
\section{Conclusion}

In this paper, we present NeRP3D, a novel point-based 3D architecture for scene reconstruction and downstream perception tasks for autonomous driving. Our approach addresses the fundamental misalignment between view transformation and neural radiance fields. Through its NeRF-resembled design, NeRP3D fully inherits NeRF's continuous representation capabilities, enabling the model to maintain consistent geometric and appearance information at arbitrary spatial locations across both scene reconstruction and open-world perceptions.
Although NeRP3D outperforms previous approaches, it struggles with depth beyond its ROI, relying on LiDAR. Additionally, its point-based architecture incurs high computational costs from adapting NeRF's output to existing detection heads. Future enhancements include temporal RGB reconstruction for consistency, density/opacity filtering for efficiency, and Gaussian splatting for real-time performance with point queries.

\section{Acknowledgement}
This work was supported by
the National Research Foundation of Korea(NRF) grant funded by the Korea government (MSIT)
and the Ministry of Trade, Industry and Resources (MOTIR) of the Republic of Korea.
(2022R1A2C200494414, RS-2025-25448249)

\bibliography{ref}

@inproceedings{caesar2020nuscenes,
  title={nuscenes: A multimodal dataset for autonomous driving},
  author={Caesar, Holger and Bankiti, Varun and Lang, Alex H and Vora, Sourabh and Liong, Venice Erin and Xu, Qiang and Krishnan, Anush and Pan, Yu and Baldan, Giancarlo and Beijbom, Oscar},
  booktitle={Proceedings of the IEEE/CVF conference on computer vision and pattern recognition},
  pages={11621--11631},
  year={2020}
}

@inproceedings{liu2022petr,
  title={Petr: Position embedding transformation for multi-view 3d object detection},
  author={Liu, Yingfei and Wang, Tiancai and Zhang, Xiangyu and Sun, Jian},
  booktitle={European conference on computer vision},
  pages={531--548},
  year={2022},
  organization={Springer}
}

@inproceedings{liu2023petrv2,
  title={Petrv2: A unified framework for 3d perception from multi-camera images},
  author={Liu, Yingfei and Yan, Junjie and Jia, Fan and Li, Shuailin and Gao, Aqi and Wang, Tiancai and Zhang, Xiangyu},
  booktitle={Proceedings of the IEEE/CVF International Conference on Computer Vision},
  pages={3262--3272},
  year={2023}
}

@inproceedings{shu20233dppe,
  title={3dppe: 3d point positional encoding for transformer-based multi-camera 3d object detection},
  author={Shu, Changyong and Deng, Jiajun and Yu, Fisher and Liu, Yifan},
  booktitle={Proceedings of the IEEE/CVF International Conference on Computer Vision},
  pages={3580--3589},
  year={2023}
}

@inproceedings{li2023bevdepth,
  title={Bevdepth: Acquisition of reliable depth for multi-view 3d object detection},
  author={Li, Yinhao and Ge, Zheng and Yu, Guanyi and Yang, Jinrong and Wang, Zengran and Shi, Yukang and Sun, Jianjian and Li, Zeming},
  booktitle={Proceedings of the AAAI conference on artificial intelligence},
  volume={37},
  number={2},
  pages={1477--1485},
  year={2023}
}

@article{li2024bevformer,
  title={Bevformer: learning bird's-eye-view representation from lidar-camera via spatiotemporal transformers},
  author={Li, Zhiqi and Wang, Wenhai and Li, Hongyang and Xie, Enze and Sima, Chonghao and Lu, Tong and Yu, Qiao and Dai, Jifeng},
  journal={IEEE Transactions on Pattern Analysis and Machine Intelligence},
  year={2024},
  publisher={IEEE}
}

@article{li2022unifying,
  title={Unifying voxel-based representation with transformer for 3d object detection},
  author={Li, Yanwei and Chen, Yilun and Qi, Xiaojuan and Li, Zeming and Sun, Jian and Jia, Jiaya},
  journal={Advances in Neural Information Processing Systems},
  volume={35},
  pages={18442--18455},
  year={2022}
}

@article{mildenhall2021nerf,
  title={Nerf: Representing scenes as neural radiance fields for view synthesis},
  author={Mildenhall, Ben and Srinivasan, Pratul P and Tancik, Matthew and Barron, Jonathan T and Ramamoorthi, Ravi and Ng, Ren},
  journal={Communications of the ACM},
  volume={65},
  number={1},
  pages={99--106},
  year={2021},
  publisher={ACM New York, NY, USA}
}

@article{wang2021neus,
  title={Neus: Learning neural implicit surfaces by volume rendering for multi-view reconstruction},
  author={Wang, Peng and Liu, Lingjie and Liu, Yuan and Theobalt, Christian and Komura, Taku and Wang, Wenping},
  journal={arXiv preprint arXiv:2106.10689},
  year={2021}
}

@inproceedings{barron2022mip,
  title={Mip-nerf 360: Unbounded anti-aliased neural radiance fields},
  author={Barron, Jonathan T and Mildenhall, Ben and Verbin, Dor and Srinivasan, Pratul P and Hedman, Peter},
  booktitle={Proceedings of the IEEE/CVF conference on computer vision and pattern recognition},
  pages={5470--5479},
  year={2022}
}

@inproceedings{barron2023zip,
  title={Zip-nerf: Anti-aliased grid-based neural radiance fields},
  author={Barron, Jonathan T and Mildenhall, Ben and Verbin, Dor and Srinivasan, Pratul P and Hedman, Peter},
  booktitle={Proceedings of the IEEE/CVF International Conference on Computer Vision},
  pages={19697--19705},
  year={2023}
}

@article{kerbl20233d,
  title={3d gaussian splatting for real-time radiance field rendering.},
  author={Kerbl, Bernhard and Kopanas, Georgios and Leimk{\"u}hler, Thomas and Drettakis, George},
  journal={ACM Trans. Graph.},
  volume={42},
  number={4},
  pages={139--1},
  year={2023}
}

@inproceedings{kirillov2023sam,
  title={Segment anything},
  author={Kirillov, Alexander and Mintun, Eric and Ravi, Nikhila and Mao, Hanzi and Rolland, Chloe and Gustafson, Laura and Xiao, Tete and Whitehead, Spencer and Berg, Alexander C and Lo, Wan-Yen and others},
  booktitle={Proceedings of the IEEE/CVF international conference on computer vision},
  pages={4015--4026},
  year={2023}
}

@article{ren2024grounded,
  title={Grounded sam: Assembling open-world models for diverse visual tasks},
  author={Ren, Tianhe and Liu, Shilong and Zeng, Ailing and Lin, Jing and Li, Kunchang and Cao, He and Chen, Jiayu and Huang, Xinyu and Chen, Yukang and Yan, Feng and others},
  journal={arXiv preprint arXiv:2401.14159},
  year={2024}
}

@article{ravi2024sam2,
  title={Sam 2: Segment anything in images and videos},
  author={Ravi, Nikhila and Gabeur, Valentin and Hu, Yuan-Ting and Hu, Ronghang and Ryali, Chaitanya and Ma, Tengyu and Khedr, Haitham and R{\"a}dle, Roman and Rolland, Chloe and Gustafson, Laura and others},
  journal={arXiv preprint arXiv:2408.00714},
  year={2024}
}

@inproceedings{yang2024unipad,
  title={Unipad: A universal pre-training paradigm for autonomous driving},
  author={Yang, Honghui and Zhang, Sha and Huang, Di and Wu, Xiaoyang and Zhu, Haoyi and He, Tong and Tang, Shixiang and Zhao, Hengshuang and Qiu, Qibo and Lin, Binbin and others},
  booktitle={Proceedings of the IEEE/CVF Conference on Computer Vision and Pattern Recognition},
  pages={15238--15250},
  year={2024}
}

@inproceedings{huang2024selfocc,
  title={Selfocc: Self-supervised vision-based 3d occupancy prediction},
  author={Huang, Yuanhui and Zheng, Wenzhao and Zhang, Borui and Zhou, Jie and Lu, Jiwen},
  booktitle={Proceedings of the IEEE/CVF Conference on Computer Vision and Pattern Recognition},
  pages={19946--19956},
  year={2024}
}

@article{zhang2023occnerf,
  title={Occnerf: Self-supervised multi-camera occupancy prediction with neural radiance fields},
  author={Zhang, Chubin and Yan, Juncheng and Wei, Yi and Li, Jiaxin and Liu, Li and Tang, Yansong and Duan, Yueqi and Lu, Jiwen},
  journal={CoRR},
  year={2023}
}

@article{xu2024gaussianpretrain,
  title={GaussianPretrain: A Simple Unified 3D Gaussian Representation for Visual Pre-training in Autonomous Driving},
  author={Xu, Shaoqing and Li, Fang and Jiang, Shengyin and Song, Ziying and Liu, Li and Yang, Zhi-xin},
  journal={arXiv preprint arXiv:2411.12452},
  year={2024}
}

@inproceedings{reading2021categorical,
  title={Categorical depth distribution network for monocular 3d object detection},
  author={Reading, Cody and Harakeh, Ali and Chae, Julia and Waslander, Steven L},
  booktitle={Proceedings of the IEEE/CVF conference on computer vision and pattern recognition},
  pages={8555--8564},
  year={2021}
}

@article{zhu2020deformable,
  title={Deformable detr: Deformable transformers for end-to-end object detection},
  author={Zhu, Xizhou and Su, Weijie and Lu, Lewei and Li, Bin and Wang, Xiaogang and Dai, Jifeng},
  journal={arXiv preprint arXiv:2010.04159},
  year={2020}
}

@inproceedings{liao2023maptr,
  title={MapTR: Structured Modeling and Learning for Online Vectorized HD Map Construction},
  author={Liao, Bencheng and Chen, Shaoyu and Wang, Xinggang and Cheng, Tianheng and Zhang, Qian and Liu, Wenyu and Huang, Chang},
  booktitle={International Conference on Learning Representations},
  year={2023}
}

@article{shin2025instagram,
  title={Instagram: Instance-level graph modeling for vectorized hd map learning},
  author={Shin, Juyeb and Jeong, Hyeonjun and Rameau, Francois and Kum, Dongsuk},
  journal={IEEE Transactions on Intelligent Transportation Systems},
  year={2025},
  publisher={IEEE}
}

@inproceedings{tong2023scene,
  title={Scene as occupancy},
  author={Tong, Wenwen and Sima, Chonghao and Wang, Tai and Chen, Li and Wu, Silei and Deng, Hanming and Gu, Yi and Lu, Lewei and Luo, Ping and Lin, Dahua and others},
  booktitle={Proceedings of the IEEE/CVF International Conference on Computer Vision},
  pages={8406--8415},
  year={2023}
}

@article{tian2023occ3d,
  title={Occ3d: A large-scale 3d occupancy prediction benchmark for autonomous driving},
  author={Tian, Xiaoyu and Jiang, Tao and Yun, Longfei and Mao, Yucheng and Yang, Huitong and Wang, Yue and Wang, Yilun and Zhao, Hang},
  journal={Advances in Neural Information Processing Systems},
  volume={36},
  pages={64318--64330},
  year={2023}
}

@inproceedings{huang2023tri,
  title={Tri-perspective view for vision-based 3d semantic occupancy prediction},
  author={Huang, Yuanhui and Zheng, Wenzhao and Zhang, Yunpeng and Zhou, Jie and Lu, Jiwen},
  booktitle={Proceedings of the IEEE/CVF conference on computer vision and pattern recognition},
  pages={9223--9232},
  year={2023}
}

@inproceedings{yang2023unisim,
  title={Unisim: A neural closed-loop sensor simulator},
  author={Yang, Ze and Chen, Yun and Wang, Jingkang and Manivasagam, Sivabalan and Ma, Wei-Chiu and Yang, Anqi Joyce and Urtasun, Raquel},
  booktitle={Proceedings of the IEEE/CVF Conference on Computer Vision and Pattern Recognition},
  pages={1389--1399},
  year={2023}
}

@inproceedings{tonderski2024neurad,
  title={Neurad: Neural rendering for autonomous driving},
  author={Tonderski, Adam and Lindstr{\"o}m, Carl and Hess, Georg and Ljungbergh, William and Svensson, Lennart and Petersson, Christoffer},
  booktitle={Proceedings of the IEEE/CVF Conference on Computer Vision and Pattern Recognition},
  pages={14895--14904},
  year={2024}
}

@article{yang2023emernerf,
  title={Emernerf: Emergent spatial-temporal scene decomposition via self-supervision},
  author={Yang, Jiawei and Ivanovic, Boris and Litany, Or and Weng, Xinshuo and Kim, Seung Wook and Li, Boyi and Che, Tong and Xu, Danfei and Fidler, Sanja and Pavone, Marco and others},
  journal={arXiv preprint arXiv:2311.02077},
  year={2023}
}

@article{wang2024distillnerf,
  title={Distillnerf: Perceiving 3d scenes from single-glance images by distilling neural fields and foundation model features},
  author={Wang, Letian and Kim, Seung Wook and Yang, Jiawei and Yu, Cunjun and Ivanovic, Boris and Waslander, Steven and Wang, Yue and Fidler, Sanja and Pavone, Marco and Karkus, Peter},
  journal={Advances in Neural Information Processing Systems},
  volume={37},
  pages={62334--62361},
  year={2024}
}

@inproceedings{cao2023scenerf,
  title={Scenerf: Self-supervised monocular 3d scene reconstruction with radiance fields},
  author={Cao, Anh-Quan and De Charette, Raoul},
  booktitle={Proceedings of the IEEE/CVF International Conference on Computer Vision},
  pages={9387--9398},
  year={2023}
}

@inproceedings{radford2021learning,
  title={Learning transferable visual models from natural language supervision},
  author={Radford, Alec and Kim, Jong Wook and Hallacy, Chris and Ramesh, Aditya and Goh, Gabriel and Agarwal, Sandhini and Sastry, Girish and Askell, Amanda and Mishkin, Pamela and Clark, Jack and others},
  booktitle={International conference on machine learning},
  pages={8748--8763},
  year={2021},
  organization={PmLR}
}

@article{oquab2023dinov2,
  title={Dinov2: Learning robust visual features without supervision},
  author={Oquab, Maxime and Darcet, Timoth{\'e}e and Moutakanni, Th{\'e}o and Vo, Huy and Szafraniec, Marc and Khalidov, Vasil and Fernandez, Pierre and Haziza, Daniel and Massa, Francisco and El-Nouby, Alaaeldin and others},
  journal={arXiv preprint arXiv:2304.07193},
  year={2023}
}

@inproceedings{hu2023planning,
  title={Planning-oriented autonomous driving},
  author={Hu, Yihan and Yang, Jiazhi and Chen, Li and Li, Keyu and Sima, Chonghao and Zhu, Xizhou and Chai, Siqi and Du, Senyao and Lin, Tianwei and Wang, Wenhai and others},
  booktitle={Proceedings of the IEEE/CVF conference on computer vision and pattern recognition},
  pages={17853--17862},
  year={2023}
}

@inproceedings{jiang2023vad,
  title={Vad: Vectorized scene representation for efficient autonomous driving},
  author={Jiang, Bo and Chen, Shaoyu and Xu, Qing and Liao, Bencheng and Chen, Jiajie and Zhou, Helong and Zhang, Qian and Liu, Wenyu and Huang, Chang and Wang, Xinggang},
  booktitle={Proceedings of the IEEE/CVF International Conference on Computer Vision},
  pages={8340--8350},
  year={2023}
}

@inproceedings{weng2024drive,
  title={Para-drive: Parallelized architecture for real-time autonomous driving},
  author={Weng, Xinshuo and Ivanovic, Boris and Wang, Yan and Wang, Yue and Pavone, Marco},
  booktitle={Proceedings of the IEEE/CVF Conference on Computer Vision and Pattern Recognition},
  pages={15449--15458},
  year={2024}
}

@inproceedings{liu2023bevfusion,
  title={Bevfusion: Multi-task multi-sensor fusion with unified bird's-eye view representation},
  author={Liu, Zhijian and Tang, Haotian and Amini, Alexander and Yang, Xinyu and Mao, Huizi and Rus, Daniela L and Han, Song},
  booktitle={2023 IEEE international conference on robotics and automation (ICRA)},
  pages={2774--2781},
  year={2023},
  organization={IEEE}
}

@inproceedings{yan2023cross,
  title={Cross modal transformer: Towards fast and robust 3d object detection},
  author={Yan, Junjie and Liu, Yingfei and Sun, Jianjian and Jia, Fan and Li, Shuailin and Wang, Tiancai and Zhang, Xiangyu},
  booktitle={Proceedings of the IEEE/CVF international conference on computer vision},
  pages={18268--18278},
  year={2023}
}

@inproceedings{kim2023crn,
  title={Crn: Camera radar net for accurate, robust, efficient 3d perception},
  author={Kim, Youngseok and Shin, Juyeb and Kim, Sanmin and Lee, In-Jae and Choi, Jun Won and Kum, Dongsuk},
  booktitle={Proceedings of the IEEE/CVF International Conference on Computer Vision},
  pages={17615--17626},
  year={2023}
}

@inproceedings{fridovich2022plenoxels,
  title={Plenoxels: Radiance fields without neural networks},
  author={Fridovich-Keil, Sara and Yu, Alex and Tancik, Matthew and Chen, Qinhong and Recht, Benjamin and Kanazawa, Angjoo},
  booktitle={Proceedings of the IEEE/CVF conference on computer vision and pattern recognition},
  pages={5501--5510},
  year={2022}
}

@article{muller2022instant,
  title={Instant neural graphics primitives with a multiresolution hash encoding},
  author={M{\"u}ller, Thomas and Evans, Alex and Schied, Christoph and Keller, Alexander},
  journal={ACM transactions on graphics (TOG)},
  volume={41},
  number={4},
  pages={1--15},
  year={2022},
  publisher={ACM New York, NY, USA}
}

@inproceedings{yu2021pixelnerf,
  title={pixelnerf: Neural radiance fields from one or few images},
  author={Yu, Alex and Ye, Vickie and Tancik, Matthew and Kanazawa, Angjoo},
  booktitle={Proceedings of the IEEE/CVF conference on computer vision and pattern recognition},
  pages={4578--4587},
  year={2021}
}

@inproceedings{chen2021mvsnerf,
  title={Mvsnerf: Fast generalizable radiance field reconstruction from multi-view stereo},
  author={Chen, Anpei and Xu, Zexiang and Zhao, Fuqiang and Zhang, Xiaoshuai and Xiang, Fanbo and Yu, Jingyi and Su, Hao},
  booktitle={Proceedings of the IEEE/CVF international conference on computer vision},
  pages={14124--14133},
  year={2021}
}

@inproceedings{wei2021nerfingmvs,
  title={Nerfingmvs: Guided optimization of neural radiance fields for indoor multi-view stereo},
  author={Wei, Yi and Liu, Shaohui and Rao, Yongming and Zhao, Wang and Lu, Jiwen and Zhou, Jie},
  booktitle={Proceedings of the IEEE/CVF international conference on computer vision},
  pages={5610--5619},
  year={2021}
}

@inproceedings{liu2022neural,
  title={Neural rays for occlusion-aware image-based rendering},
  author={Liu, Yuan and Peng, Sida and Liu, Lingjie and Wang, Qianqian and Wang, Peng and Theobalt, Christian and Zhou, Xiaowei and Wang, Wenping},
  booktitle={Proceedings of the IEEE/CVF Conference on Computer Vision and Pattern Recognition},
  pages={7824--7833},
  year={2022}
}

@inproceedings{roessle2022dense,
  title={Dense depth priors for neural radiance fields from sparse input views},
  author={Roessle, Barbara and Barron, Jonathan T and Mildenhall, Ben and Srinivasan, Pratul P and Nie{\ss}ner, Matthias},
  booktitle={Proceedings of the IEEE/CVF Conference on Computer Vision and Pattern Recognition},
  pages={12892--12901},
  year={2022}
}

@inproceedings{deng2022depth,
  title={Depth-supervised nerf: Fewer views and faster training for free},
  author={Deng, Kangle and Liu, Andrew and Zhu, Jun-Yan and Ramanan, Deva},
  booktitle={Proceedings of the IEEE/CVF conference on computer vision and pattern recognition},
  pages={12882--12891},
  year={2022}
}

@article{wei2023depth,
  title={Depth-guided optimization of neural radiance fields for indoor multi-view stereo},
  author={Wei, Yi and Liu, Shaohui and Zhou, Jie and Lu, Jiwen},
  journal={IEEE Transactions on Pattern Analysis and Machine Intelligence},
  volume={45},
  number={9},
  pages={10835--10849},
  year={2023},
  publisher={IEEE}
}

@inproceedings{godard2017unsupervised,
  title={Unsupervised monocular depth estimation with left-right consistency},
  author={Godard, Cl{\'e}ment and Mac Aodha, Oisin and Brostow, Gabriel J},
  booktitle={Proceedings of the IEEE conference on computer vision and pattern recognition},
  pages={270--279},
  year={2017}
}

@inproceedings{godard2019digging,
  title={Digging into self-supervised monocular depth estimation},
  author={Godard, Cl{\'e}ment and Mac Aodha, Oisin and Firman, Michael and Brostow, Gabriel J},
  booktitle={Proceedings of the IEEE/CVF international conference on computer vision},
  pages={3828--3838},
  year={2019}
}

@inproceedings{zhou2017unsupervised,
  title={Unsupervised learning of depth and ego-motion from video},
  author={Zhou, Tinghui and Brown, Matthew and Snavely, Noah and Lowe, David G},
  booktitle={Proceedings of the IEEE conference on computer vision and pattern recognition},
  pages={1851--1858},
  year={2017}
}

@inproceedings{caron2021emerging,
  title={Emerging properties in self-supervised vision transformers},
  author={Caron, Mathilde and Touvron, Hugo and Misra, Ishan and J{\'e}gou, Herv{\'e} and Mairal, Julien and Bojanowski, Piotr and Joulin, Armand},
  booktitle={Proceedings of the IEEE/CVF international conference on computer vision},
  pages={9650--9660},
  year={2021}
}

@inproceedings{yang2023bevformer,
  title={Bevformer v2: Adapting modern image backbones to bird's-eye-view recognition via perspective supervision},
  author={Yang, Chenyu and Chen, Yuntao and Tian, Hao and Tao, Chenxin and Zhu, Xizhou and Zhang, Zhaoxiang and Huang, Gao and Li, Hongyang and Qiao, Yu and Lu, Lewei and others},
  booktitle={Proceedings of the IEEE/CVF Conference on Computer Vision and Pattern Recognition},
  pages={17830--17839},
  year={2023}
}

@inproceedings{wei2023surroundocc,
  title={Surroundocc: Multi-camera 3d occupancy prediction for autonomous driving},
  author={Wei, Yi and Zhao, Linqing and Zheng, Wenzhao and Zhu, Zheng and Zhou, Jie and Lu, Jiwen},
  booktitle={Proceedings of the IEEE/CVF International Conference on Computer Vision},
  pages={21729--21740},
  year={2023}
}

@misc{mmdet3d2020,
    title={{MMDetection3D: OpenMMLab} next-generation platform for general {3D} object detection},
    author={MMDetection3D Contributors},
    howpublished = {\url{https://github.com/open-mmlab/mmdetection3d}},
    year={2020}
}

@inproceedings{zhang2025visionpad,
  title={Visionpad: A vision-centric pre-training paradigm for autonomous driving},
  author={Zhang, Haiming and Zhou, Wending and Zhu, Yiyao and Yan, Xu and Gao, Jiantao and Bai, Dongfeng and Cai, Yingjie and Liu, Bingbing and Cui, Shuguang and Li, Zhen},
  booktitle={Proceedings of the Computer Vision and Pattern Recognition Conference},
  pages={17165--17175},
  year={2025}
}

@article{wilson2023argoverse,
  title={Argoverse 2: Next generation datasets for self-driving perception and forecasting},
  author={Wilson, Benjamin and Qi, William and Agarwal, Tanmay and Lambert, John and Singh, Jagjeet and Khandelwal, Siddhesh and Pan, Bowen and Kumar, Ratnesh and Hartnett, Andrew and Pontes, Jhony Kaesemodel and others},
  journal={arXiv preprint arXiv:2301.00493},
  year={2023}
}
\bibliographystyle{iclr2026/iclr2026_conference}

\newpage
\appendix

\begin{table*}[t]
	\centering
    \vspace{-10pt}
    \caption{
        \textbf{3D object detection}
        on the nuScenes \textit{val} set. 
		$\dagger$ denotes the result evaluated on input resolutions of $800\times450$ using MMDetection3D~\citep{mmdet3d2020} by integrating UVTR detection head~\citep{li2022unifying, yang2024unipad}. The other results are based on $1600\times900$ input resolution.
    }
    \resizebox{\linewidth}{!}{
    \begin{tabular}{l|c|cc|c|c|c|c|c}
        \toprule
        Method & Pre-train & NDS$\uparrow$ & mAP$\uparrow$ & mATE$\downarrow$ & mASE$\downarrow$ & mAOE$\downarrow$ & mAVE$\downarrow$ & mAAE$\downarrow$ \\
        \midrule
        BEVFormer-S & ImageNet & 44.8 & 37.5 & -     & -     & -     & -     & -     \\
        UVTR-C      & ImageNet & 44.1 & 37.2 & 0.735 & 0.269 & 0.397 & 0.761 & 0.193 \\
        PETR        & ImageNet & 44.2 & 37.0 & 0.711 & 2.670 & 0.383 & 0.865 & 0.201 \\
		3DPPE       & ImageNet & 45.8 & 39.1 & -     & -     & -     & -     & -     \\
        BEVFormerV2 & ImageNet & 46.7 & 39.6 & 0.709 & 0.274 & 0.368 & 0.768 & 0.196 \\
        CMT-C       & ImageNet & 46.0 & 40.6 & -     & -     & -     & -     & -     \\
        \midrule
        TPVFormer$^\dagger$       & SelfOcc & 33.5 & 31.0 & 0.785 & 0.285 & 0.729 & 1.232 & 0.399 \\
        UVTR-C$^\dagger$          & UniPAD  & 37.1 & 33.7 & 0.734 & 0.283 & 0.603 & 1.250 & 0.359 \\
        \textbf{NeRP3D}$^\dagger$ & Ours    & \textbf{39.2} & \textbf{35.8} & 0.719 & 0.288 & 0.640 & 0.977 & 0.250 \\
        \midrule
        UVTR-C          & UniPAD           & 45.5 & 41.6 & 0.674 & 0.277 & 0.418 & 0.930 & 0.234 \\
        UVTR-C          & GaussianPretrain & 47.2 & 41.7 & 0.676 & 0.278 & 0.394 & 0.815 & 0.200 \\
        \textbf{NeRP3D} & Ours & \textbf{47.3} & \textbf{42.8} & 0.664 & 0.276 & 0.425 & 0.811 & 0.196 \\
		\bottomrule
	\end{tabular}
    }
\label{tab:det}
\vspace{-5pt}
\end{table*}
\definecolor{nbarrier}{RGB}{255, 120, 50}
\definecolor{nbicycle}{RGB}{255, 192, 203}
\definecolor{nbus}{RGB}{255, 255, 0}
\definecolor{ncar}{RGB}{0, 150, 245}
\definecolor{nconstruct}{RGB}{0, 255, 255}
\definecolor{nmotor}{RGB}{200, 180, 0}
\definecolor{npedestrian}{RGB}{255, 0, 255}
\definecolor{ntraffic}{RGB}{255, 240, 150}
\definecolor{ntrailer}{RGB}{135, 60, 0}
\definecolor{ntruck}{RGB}{255, 0, 0}
\definecolor{ndriveable}{RGB}{213, 213, 213}
\definecolor{nother}{RGB}{139, 137, 137}
\definecolor{nsidewalk}{RGB}{75, 0, 75}
\definecolor{nterrain}{RGB}{150, 240, 80}
\definecolor{nmanmade}{RGB}{160, 32, 240}
\definecolor{nvegetation}{RGB}{0, 175, 0}
\definecolor{nothers}{RGB}{0, 0, 0}

\begin{table*}[t]
    \footnotesize
    \setlength{\tabcolsep}{0.0025\linewidth}
    \centering
    \caption{
        \textbf{3D occupancy prediction}.
        We compare our method against \textit{state-of-the-art} occupancy prediction approaches on the Occ3d-nuScenes \textit{val} set.
        Results for BEVDet, BEVFormer, TPVFormer, and CTF-Occ are directly taken from Occ3d~\citep{tian2023occ3d}.
        $\dagger$ denotes the result reproduced using MMDetection3D~\citep{mmdet3d2020} on input resolutions of $800\times450$.
        $\ast$ denotes that the result is directly taken from VisionPAD~\citep{zhang2025visionpad}, which is pre-trained only with camera modality and evaluated on input resolutions of $1600\times900$.
    }
    \resizebox{0.9\linewidth}{!}{
    \begin{tabular}{l|c| r r r r r r r r r r r r r r r r r}
        \toprule
        Method
        & mIoU
        & \multicolumn{1}{c}{\rotatebox{90}{\textcolor{ncar}{$\blacksquare$} car}}
        & \multicolumn{1}{c}{\rotatebox{90}{\textcolor{nbus}{$\blacksquare$} bus}}
        & \multicolumn{1}{c}{\rotatebox{90}{\textcolor{nbicycle}{$\blacksquare$} bicycle}}
        & \multicolumn{1}{c}{\rotatebox{90}{\textcolor{nbarrier}{$\blacksquare$} barrier}}
        & \multicolumn{1}{c}{\rotatebox{90}{\textcolor{nvegetation}{$\blacksquare$} vegetation}}
        & \multicolumn{1}{c}{\rotatebox{90}{\textcolor{nconstruct}{$\blacksquare$} const. veh.}}
        & \multicolumn{1}{c}{\rotatebox{90}{\textcolor{nmotor}{$\blacksquare$} motorcycle}}
        & \multicolumn{1}{c}{\rotatebox{90}{\textcolor{npedestrian}{$\blacksquare$} pedestrian}}
        & \multicolumn{1}{c}{\rotatebox{90}{\textcolor{ntraffic}{$\blacksquare$} traffic cone}}
        & \multicolumn{1}{c}{\rotatebox{90}{\textcolor{ntrailer}{$\blacksquare$} trailer}}
        & \multicolumn{1}{c}{\rotatebox{90}{\textcolor{ntruck}{$\blacksquare$} truck}}
        & \multicolumn{1}{c}{\rotatebox{90}{\textcolor{ndriveable}{$\blacksquare$} drive. suf.}}
        & \multicolumn{1}{c}{\rotatebox{90}{\textcolor{nother}{$\blacksquare$} other flat}}
        & \multicolumn{1}{c}{\rotatebox{90}{\textcolor{nsidewalk}{$\blacksquare$} sidewalk}}
        & \multicolumn{1}{c}{\rotatebox{90}{\textcolor{nterrain}{$\blacksquare$} terrain}}
        & \multicolumn{1}{c}{\rotatebox{90}{\textcolor{nmanmade}{$\blacksquare$} manmade}}
        & \multicolumn{1}{c}{\rotatebox{90}{\textcolor{nothers}{$\blacksquare$} others}}\\
        \midrule
        BEVDet    & 19.4 & 34.5 & 32.3 &  0.2 & 30.3 & 15.1 & 13.0 & 10.3 & 10.4 &  6.3 &  8.9 & 23.7 & 52.7 & 24.6 & 26.1 & 22.3 & 15.0 & 4.4 \\
        BEVFormer & 26.9 & 42.4 & 40.4 & 17.9 & 37.8 & 17.7 &  7.4 & 23.9 & 21.8 & 21.0 & 22.4 & 30.7 & 55.4 & 28.4 & 36.0 & 28.1 & 20.0 & 5.9 \\
        TPVFormer & 27.8 & 45.9 & 40.8 & 13.7 & 38.9 & 16.8 & 17.2 & 20.0 & 18.9 & 14.3 & 26.7 & 34.2 & 55.7 & 35.5 & 37.6 & 30.7 & 19.4 & 7.2 \\
        CTF-Occ   & 28.5 & 42.2 & 38.3 & 20.6 & 39.3 & 18.0 & 16.9 & 24.5 & 22.7 & 21.1 & 23.0 & 31.1 & 53.3 & 33.8 & 38.0 & 33.2 & 20.8 & 8.1 \\
        \midrule
        SelfOcc$^\dagger$         & 29.7          & 43.8 & 40.0 & 10.0 & 36.3 & 30.6 & 13.7 & 11.8 & 16.5 & 15.7 & 23.2 & 29.3 & 79.1 & 37.3 & 47.7 & 28.0 & 34.8 & 6.2 \\
        UniPAD$^\dagger$          & 34.1          & 45.8 & 42.3 & 13.0 & 39.7 & 38.1 & 19.4 & 14.3 & 20.0 & 17.7 & 27.4 & 33.1 & 80.0 & 38.7 & 49.4 & 50.6 & 42.8 & 6.5 \\
        VisionPAD$^\ast$ & 35.4 & \multicolumn{1}{c}{-} & \multicolumn{1}{c}{-} & \multicolumn{1}{c}{-} & \multicolumn{1}{c}{-} & \multicolumn{1}{c}{-} & \multicolumn{1}{c}{-} & \multicolumn{1}{c}{-} & \multicolumn{1}{c}{-} & \multicolumn{1}{c}{-} & \multicolumn{1}{c}{-} & \multicolumn{1}{c}{-} & \multicolumn{1}{c}{-} & \multicolumn{1}{c}{-} & \multicolumn{1}{c}{-} & \multicolumn{1}{c}{-} & \multicolumn{1}{c}{-} & \multicolumn{1}{c}{-} \\
        \textbf{NeRP3D}$^\dagger$ & \textbf{35.5} & 49.4 & 43.9 & 15.0 & 41.0 & 38.8 & 19.2 & 20.0 & 23.6 & 16.5 & 27.9 & 36.7 & 81.0 & 37.4 & 49.8 & 53.6 & 43.9 & 5.5 \\
        \bottomrule
    \end{tabular}
    }
\label{tab:occ}
\vspace{-5pt}
\end{table*}

\section{Downstream Detection Tasks}
\label{sup_a}
A detailed analysis of NeRP3D's performance is provided on three downstream perception tasks: 3D object detection, 3D occupancy prediction, and HD map construction. We expand upon the results presented in Sec.~\ref{main_results} and Tab.~\ref{tab:downstream}, with a focus on comprehensive comparisons against state-of-the-art methods, including those leveraging 3DGS (3D Gaussian Splatting)-based pre-training.

\begin{figure*}[t]
    \centering
    \includegraphics[width=0.93\linewidth]{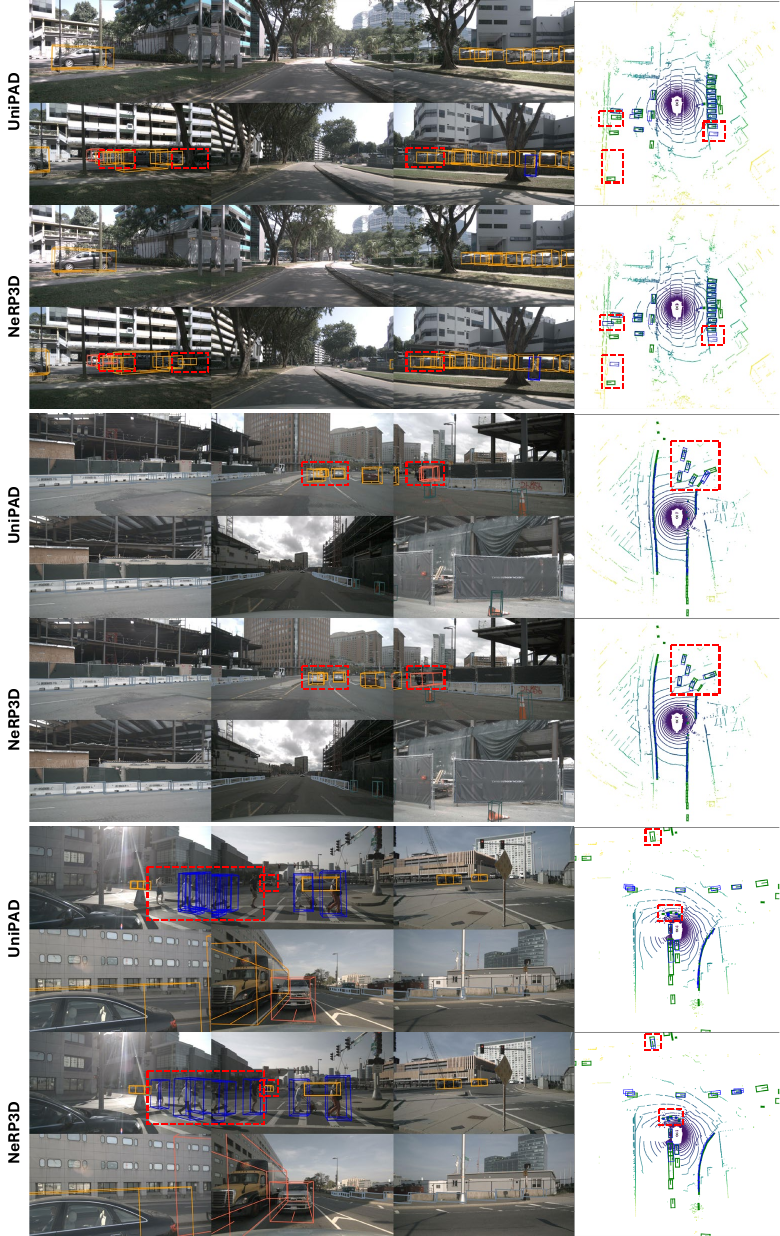}
    \vspace{-5pt}
    \caption{
        Qualitative comparison of 3D object detection results.
        NeRP3D consistently generates more accurate and reliable 3D bounding boxes.
        It demonstrates key advantages such as successfully detecting partially occluded objects in dense crowds (top row), reducing false positives for cleaner predictions (middle row), and more accurately localizing the position of small objects like pedestrians (bottom row).
    }
    \label{fig8}
\end{figure*}

As shown in Tab.~\ref{tab:det}, NeRP3D achieves state-of-the-art performance in 3D object detection among NeRF-based pre-training methods, with an NDS of 47.3 and an mAP of 42.8. This represents a significant improvement over UniPAD, with gains of 1.8 NDS and 1.2 mAP when both are fine-tuned on the UVTR-C detector.
Crucially, NeRP3D also outperforms GaussianPretrain~\citep{xu2024gaussianpretrain}, which still relies on a view transformation backbone. In comparison, NeRP3D achieves a higher NDS (47.3 vs. 47.2) and a more substantial lead in mAP (42.8 vs. 41.7).
The enhanced performance is attributed to NeRP3D's fine-grained 3D representation, which provides the necessary detail to identify far or occluded targets and resolve individuals within dense crowds, as shown in Fig.~\ref{fig8}

\begin{figure*}[t]
    \centering
    \includegraphics[width=1.0\linewidth]{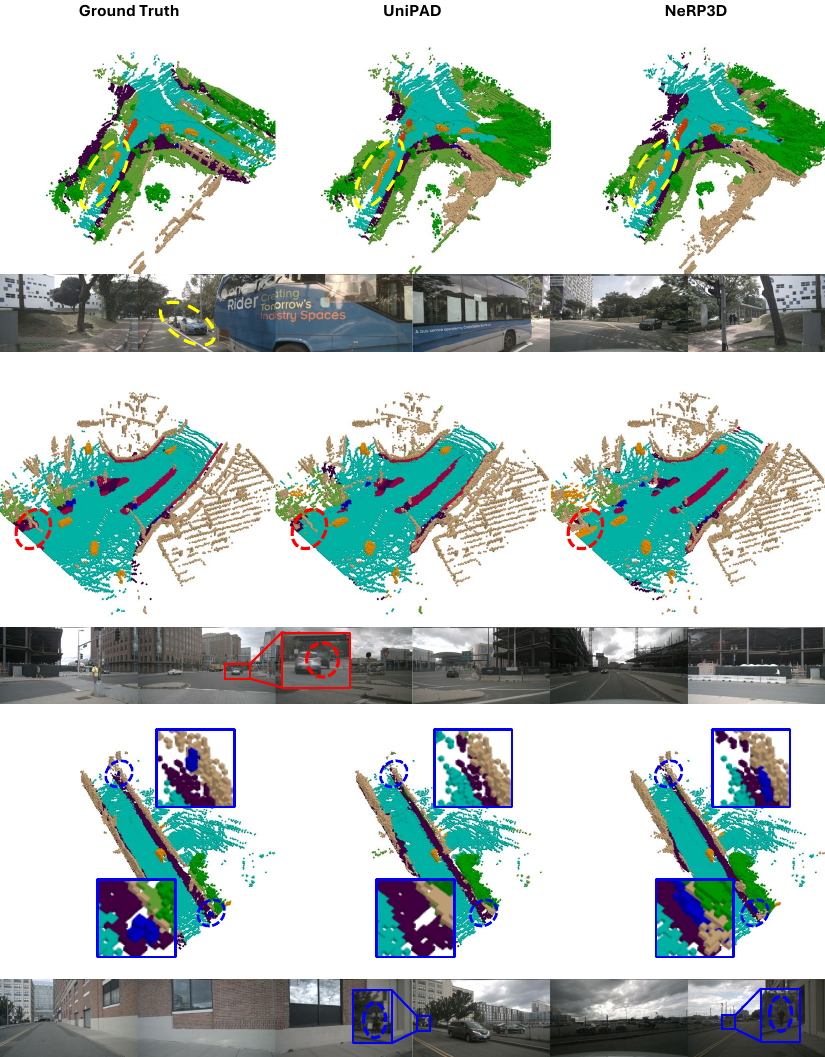}
    \vspace{-10pt}
    \caption{
        Qualitative comparison of occupancy prediction results.
        NeRP3D produces more detailed and complete occupancy predictions compared to UniPAD. NeRP3D excels at distinguishing individual objects that are close together, as shown by its clear separation of the vehicles (top row, yellow). Furthermore, it demonstrates a superior ability to detect objects that are entirely missed in the ground truth annotation, likely due to occlusion (middle row, red). The robust perception ability of NeRP3D also extends to resolving smaller, distant objects, such as pedestrians (bottom row, blue), contributing to more accurate and reliable scene understanding.
    }
    \label{fig9}
\end{figure*}

For 3D occupancy prediction, NeRP3D's ability to model continuous geometry and appearance translates into superior performance. As demonstrated in Tab.~\ref{tab:occ}, our method achieves an mIoU of 35.5, surpassing both UniPAD (34.1 mIoU) and SelfOcc (29.7 mIoU) by a significant margin.
We further compare NeRP3D with VisionPAD~\citep{zhang2025visionpad}, a vision-centric pre-training also based on 3D Gaussians. Even though VisionPAD is pre-trained only with camera modality, but evaluated on the higher resolution $1600\times900$, NeRP3D achieves a competitive overall mIoU (35.5 vs. 35.4). A class-level breakdown reveals that NeRP3D shows notable improvements in thin and small categories, as shown in Fig.~\ref{fig9}, such as bicycle (15.0 vs. 13.0), motorcycle (20.0 vs. 14.3), and pedestrian (23.6 vs. 20.0).

The comprehensive results for downstream perception tasks indicate that our NeRP3D, which avoids the conflicting priors between the pre-training method and 3D backbone, enables the learning of continuous and fine-grained 3D representations that directly benefit downstream detection tasks.

\begin{table*}[!t]
    \centering
    \vspace{-10pt}
    \caption{
        \textbf{HD map construction}
        on the nuScenes \textit{val} set.
        ``C'' and ``L'' denote camera and LiDAR modalities, respectively. Results for HDMapNet and VectorMapNet are directly taken from MapTR\citep{liao2023maptr}.
    }
    \resizebox{0.9\textwidth}{!}{%
    \begin{tabular}{l|c|c|c|cccc}
        \toprule
        Method & Modality & Pre-train & Epochs & mAP & AP$_{ped}$ & AP$_{divider}$ & AP$_{boundary}$ \\
        \toprule
        HDMapNet        & C      & ImageNet &  30 & 23.0 & 14.4 & 21.7 & 33.0 \\
        HDMapNet        & L      & ImageNet &  30 & 24.1 & 10.4 & 24.1 & 37.9 \\
        HDMapNet        & C \& L & ImageNet &  30 & 31.0 & 16.3 & 29.6 & 46.7 \\
        \midrule
        VectorMapNet    & C      & ImageNet & 110 & 40.9 & 36.1 & 47.3 & 39.3 \\
        VectorMapNet    & L      & ImageNet & 110 & 34.0 & 25.7 & 37.6 & 38.6 \\
        VectorMapNet    & C \& L & ImageNet & 110 & 45.2 & 37.6 & 50.5 & 47.5 \\
        \midrule
        MapTR-tiny      & C      & ImageNet &  24 & 49.9 & 52.0 & 45.3 & 52.4 \\
        \midrule
        TPVFormer       & C & SelfOcc & 24 & 53.9 & 47.8 & 55.6 & 58.3 \\
        UVTR-C          & C & UniPAD  & 24 & 57.8 & \textbf{54.8} & 58.5 & 61.5 \\
        \textbf{NeRP3D} & C & Ours    & 24 & \textbf{59.1} & 52.9 & \textbf{62.2} & \textbf{62.2} \\
        \bottomrule
    \end{tabular}
    }
\label{tab:map}
\vspace{-5pt}
\end{table*}
\begin{table*}[!t]
    \centering
    \caption{
        \textbf{Depth estimation} on nuScenes \textit{val} set.
        We conduct evaluation at a downsampled resolution of $114\times64$ for EmerNeRF~\citep{yang2023emernerf} and DistillNeRF~\citep{wang2024distillnerf} and $200\times112$ for others.
        $\dagger$ denotes per-scene optimization, not feedforward model.
        $\ast$ denotes only \textit{depth-optimized} variant of SelfOcc~\citep{huang2024selfocc}.
        The results of EmerNeRF and DistillNeRF are taken from the paper of DistillNeRF.
    }
    \resizebox{1.0\textwidth}{!}{%
        \begin{tabular}{l|cccc|ccc}
        \toprule
        Method & Abs Rel$\downarrow$ & Sq Rel$\downarrow$ & RMSE$\downarrow$ & RMSE log $\downarrow$ & $\delta<1.25\uparrow$ & $\delta<1.25^2\uparrow$ & $\delta<1.25^3\uparrow$ \\
        \midrule
        EmerNeRF$^\dagger$ & 0.073 & 0.346 & 2.696 & 0.159 & 0.942 & 0.975 & 0.986 \\
        \midrule
        DistillNeRF    & 0.248 & 3.090 & 6.096 & 0.312 & 0.704 & \underline{0.885} & \underline{0.947} \\
        DistillNeRF-D  & 0.233 & 2.890 & \underline{5.890} & \underline{0.296} & 0.703 & 0.881 & 0.945 \\
        DistillNeRF-DV & 0.223 & \textbf{1.776} & \textbf{5.461} & \textbf{0.293} & \underline{0.763} & \textbf{0.903} & \textbf{0.961} \\
        \midrule
        SelfOcc        & 0.311 & 3.808 & 8.503 & 0.391 & 0.641 & 0.803 & 0.888 \\
        SelfOcc$^\ast$ & 0.215 & 2.743 & 6.706 & 0.316 & 0.753 & 0.875 & 0.932 \\
        UniPAD         & \underline{0.218} & 2.512 & 7.937 & 0.356 & \underline{0.763} & 0.869 & 0.921 \\
        NeRP3D         & \textbf{0.183} & \underline{2.274} & 7.884 & 0.353 & \textbf{0.799} & 0.883 & 0.926 \\
        \bottomrule
        \end{tabular}
    }
\label{tab:depth}
\vspace{-5pt}
\end{table*}

\begin{SCtable}
    \centering
    \caption{
        \textbf{RGB reconstruction}
        on nuScenes \textit{val} set at a resolution of $228\times114$ for EmerNeRF~\citep{yang2023emernerf} and DistillNeRF~\citep{wang2024distillnerf} and $400\times225$ for others. The results of EmerNeRF and DistillNeRF are taken from the paper of DistillNeRF.
    }
    \resizebox{0.48\linewidth}{!}{
        \begin{tabular}{lccc}
        \toprule
        Method          & PSNR $\uparrow$ & SSIM $\uparrow$ & LPIPS $\downarrow$ \\
        \midrule
        EmerNeRF        & 30.88           & 0.879           & - \\
        DistillNeRF-D   & 30.11           & 0.917           & - \\
        SelfOcc         & 18.82           & 0.536           & 0.657              \\
        UniPAD          & 21.14           & 0.549           & 0.634              \\
        \textbf{NeRP3D} & \textbf{33.42}  & \textbf{0.969}  & \textbf{0.070}     \\
        \bottomrule
        \end{tabular}
    }
\label{tab:rgb}
\vspace{-10pt}
\end{SCtable}

\section{Pretext Scene Reconstruction Tasks}
\label{sup_b}

The overall performance of RGB reconstruction and depth estimation is compared with previous NeRF-based pre-training methods~\citep{yang2024unipad, huang2024selfocc} and comparable methods on the nuScenes \textit{val} set, as shown in Tab.~\ref{tab:depth} and \ref{tab:rgb}. Specifically, EmerNeRF~\citep{yang2023emernerf} is a \textit{per-scene} optimization model, and the variants of DistillNeRF~\citep{wang2024distillnerf} are \textit{without} distillation, \textit{with} depth distillation (noted as ``D''), and \textit{with} virtual camera distillation (noted as ``V'').

The depth estimation results in Tab.~\ref {tab:depth} demonstrate clear benefits from our NeRF-inherited representation learning.
SelfOcc$^\ast$ shows competitive depth estimation, but this variant does not support RGB reconstruction. On the other hand, the variant of SelfOcc that supports both RGB and depth reconstruction exhibits comparatively lower accuracy.
Compared to UniPAD, our method achieves better performance across multiple metrics, such as AbsRel (0.183 vs. 0.218), SqRel (2.274 vs. 2.512), and RMSE (7.884 vs. 7.937). Moreover, accuracy within specific depth thresholds ($\delta$ metrics) further underscores the robustness of our model in reconstructing precise depth values.
When compared with DistillNeRF, which is specifically designed for scene reconstruction, our NeRP3D achieves competitive depth estimation accuracy despite not relying on dense depth maps obtained from per-scene optimization~\citep{yang2023emernerf} or distillation from 2D foundation models~\citep{radford2021learning, oquab2023dinov2}.

For RGB reconstruction, NeRP3D significantly outperforms previous approaches, as shown in Tab.~\ref{tab:rgb}. Compared to previous feedforward methods and EmerNeRF, PSNR and SSIM are improved by 33.42 and 0.969, respectively. Our method also notably reduces LPIPS, reflecting more perceptually accurate reconstructions over UniPAD and SelfOcc by 0.070.

To quantitatively verify the conflicting prior between discrete view transformation and continuous neural rendering representations, we evaluated reconstruction performance across varying resolutions as shown in Tab.~\ref{tab:conflict}. The view transformation method (UniPAD~\citep{yang2024unipad}) degrades at higher resolutions, confirming that discrete voxel grids act as a "low-pass filter". As a result, UniPAD masks errors at coarse scales but fails to capture high-frequency details due to the rigid bottleneck. In contrast, NeRP3D demonstrates superior representational fidelity with a widening performance gap at finer scales. This quantitatively proves that our continuous architecture resolves the structural mismatch, successfully modeling fine-grained geometry that fixed grids cannot capture.

\begin{table}[tb]
    \centering
    \vspace{-10pt}
    \caption{\textcolor{black}{
        \textbf{Multi-resolution reconstruction analysis.}
        We evaluate reconstruction quality across varying image scales (from $1/16$ to $1/4$) to isolate the impact of discretization levels on representational fidelity.
    }}
    \resizebox{0.60\linewidth}{!}{
        \begin{tabular}{l|l|ccc}
        \toprule
        Method                                        & Scale & PSNR$\uparrow$  & SSIM$\uparrow$  & LPIPS$\downarrow$ \\
        \midrule
        \multirow{3}{*}{UniPAD~\citep{yang2024unipad}} & 1/16  & 23.55           & 0.796           & 0.250 \\
                                                      & 1/8   & 22.49           & 0.664           & 0.442 \\
                                                      & 1/4   & 21.14           & 0.549           & 0.634 \\
        \midrule
        \multirow{3}{*}{\textbf{NeRP3D}}              & 1/16  & 26.00           & 0.862           & 0.116 \\
                                                      & 1/8   & 29.40           & 0.919           & 0.090 \\
                                                      & 1/4   & 33.42           & 0.969           & 0.070 \\
        \bottomrule
        \end{tabular}
    }
    \label{tab:conflict}
\end{table}

\section{Cross task generalization}
\label{sup_c}

\begin{table}[tb]
\centering
\vspace{-10pt}
\caption{\textcolor{black}{
    \textbf{Evaluation of Cross-Task Generalization and Structural Retention.}
    (a) Per-pixel evaluation: Standard metrics measuring absolute reconstruction errors, which are sensitive to scale shifts. (b) Structural evaluation: Scale-invariant and perceptual metrics assessing geometric fidelity and structural integrity, independent of feature scale variations induced during fine-tuning.
}}
\vspace{10pt}
    \begin{minipage}{\textwidth}
    \centering
    (a) \textbf{Per-pixel evaluation} \\
    \resizebox{0.85\linewidth}{!}{%
        \begin{tabular}{l|cccc|cc}
        \toprule
        Method                       & Abs Rel$\downarrow$ & Sq Rel$\downarrow$ & RMSE$\downarrow$ & RMSE log$\downarrow$ 
                                     & PSNR$\uparrow$      & SSIM$\uparrow$\\
        \midrule
        UniPAD~\citep{yang2024unipad} & 0.477               & 6.914              & 15.104           & 1.056
                                     & 11.623              & 0.283 \\
        \textbf{NeRP3D}              & 2.192               & 12.372             & 19.459           & 1.185
                                     &  9.308              & 0.135 \\
        \bottomrule
        \end{tabular}
    }
    \end{minipage}

    \vspace{10pt}
    \begin{minipage}{\textwidth}
    \centering
    (b) \textbf{Structural and scale-invariant evaluation} \\
    \resizebox{0.95\linewidth}{!}{%
        \begin{tabular}{l|cc|cccc}
        \toprule
        Method                       & SI RMSE$\downarrow$ & Grad Loss$\downarrow$ & GMSD$\downarrow$ & LPIPS$\downarrow$ 
                                     & PSNR-HM$\uparrow$   & SSIM-HM$\uparrow$\\
        \midrule
        UniPAD~\citep{yang2024unipad} & 0.859               & 90.164                & 0.306            & 0.863
                                     & 12.319              & \textbf{0.300} \\
        \textbf{NeRP3D}              & \textbf{0.643}      & \textbf{83.739}       & \textbf{0.289}   & \textbf{0.671}
                                     & \textbf{12.839}     & 0.285 \\
        \bottomrule
        \end{tabular}
    }
    \end{minipage}
    \vspace{-10pt}
\label{tab:cross}
\end{table}

We investigate whether the learned 3D representations remain valid across different task objectives, specifically evaluating the "Radiance Modeling Ability" of the backbone after fine-tuning for occupancy prediction. In this experiment, we utilize the backbone encoder fine-tuned for the downstream task while keeping the pre-trained rendering heads (RGB and SDF decoder) frozen.

As shown in Fig.~\ref{fig11}, there is a stark contrast in the retained representations; the view transformation method like UniPAD~\citep{yang2024unipad} suffers from catastrophic forgetting, producing blurry outputs that indicate a loss of fine-grained 3D information and a collapse into mean regression. In contrast, NeRP3D successfully retains structural understanding, with key elements remaining clearly distinguishable. Quantitative results in Tab.~\ref{tab:cross} further support this.
While standard per-pixel metrics are sensitive to feature scale shifts induced during fine-tuning, often favoring the mean regression of UniPAD, NeRP3D significantly outperforms the view transformation method in scale-invariant (SI-RMSE) and perceptual (LPIPS, GMSD) metrics. This confirms that, unlike view transformation methods that overfit to specific tasks and collapse into mean regression, NeRP3D learns a robust and continuous representation that maintains geometric fidelity across diverse downstream objectives.

\newpage

\newpage

\begin{figure*}[!t]
    \centering
    \includegraphics[width=1.0\linewidth]{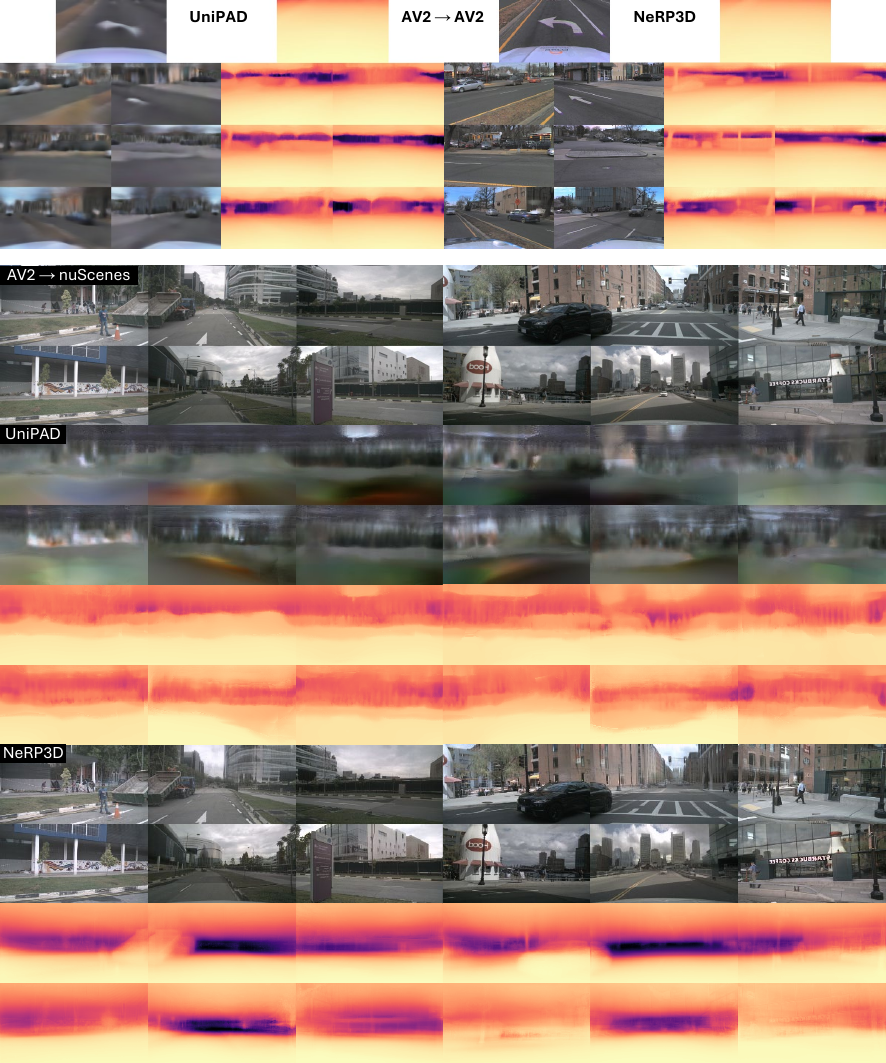}
    \vspace{-10pt}
    \caption{\textcolor{black}{
        Qualitative comparison of zero-shot transfer for reconstruction from Argoverse 2 to nuScenes dataset (pre-training phase).
        Models were pre-trained on Argoverse 2 (AV2) and evaluated without any fine-tuning on the target domain (nuScenes).
        (Top: AV2 $\rightarrow$ AV2) In-domain reconstruction results. Both models demonstrate that pre-training on AV2 was successful, reconstructing scene details within the source domain.
        (Bottom: AV2 $\rightarrow$ nuScenes) Zero-shot transfer results to nuScenes. When applying AV2-trained weights directly to the distinct camera geometry of nuScenes, UniPAD fails to render meaningful structure (blurry artifacts), revealing the vulnerability of fixed voxel grids to sensor layout changes. In contrast, NeRP3D maintains high-fidelity rendering, demonstrating that its point-based architecture is sensor-agnostic and robust to extrinsic shifts.
    }}
    \label{fig10}
\end{figure*}

\begin{figure*}[!t]
    \centering
    \includegraphics[width=1.0\linewidth]{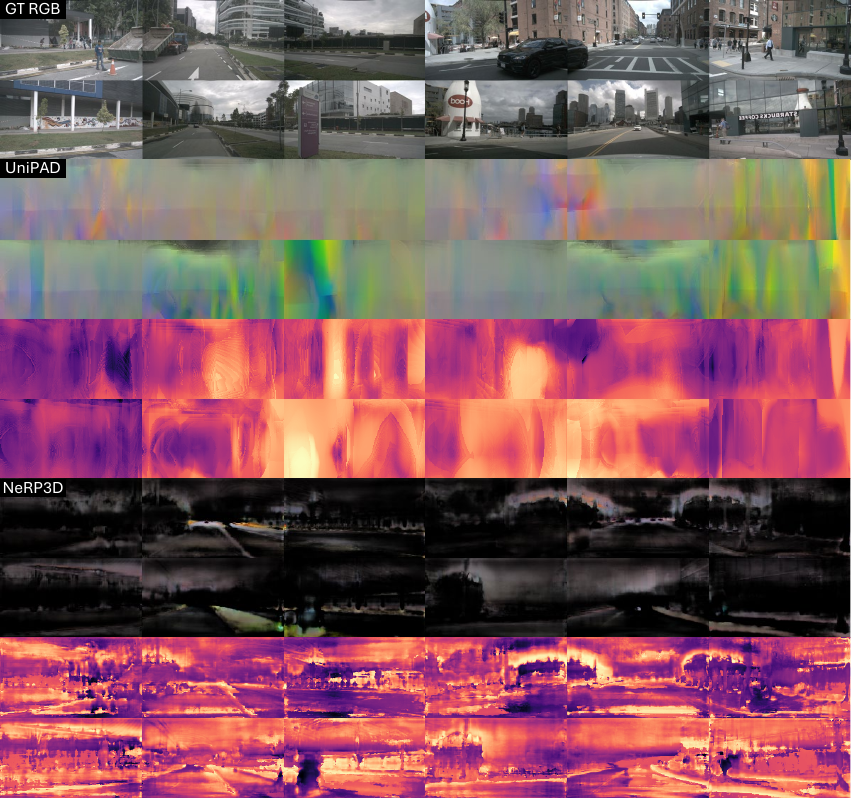}
    \vspace{-10pt}
    \caption{\textcolor{black}{
        Qualitative evaluation of radiance modeling ability after fine-tuning for occupancy prediction.
        We visualize rendering results using backbones fine-tuned for the occupancy task, with pre-trained decoders frozen. (Row 2-3) UniPAD suffers from catastrophic forgetting, producing "blurry gray" outputs. The model loses 3D structural information and resorts to mean regression to minimize loss. (Row 4-5) NeRP3D successfully retains structural understanding despite the task shift. Key elements like vehicles, road boundaries, and building structures remain clearly distinguishable in both RGB and Depth renderings. This proves that NeRP3D learns a task-agnostic continuous representation that remains valid across different downstream objectives.
    }}
    \label{fig11}
\end{figure*}

\end{document}